\documentclass{article}

\usepackage{microtype}
\usepackage{graphicx}
\usepackage{subcaption}
\usepackage{booktabs} % for professional tables
\usepackage[table]{xcolor}
\usepackage{amssymb}
\usepackage{pifont}
\newcommand{\cmark}{\ding{51}} % ✓
\newcommand{\xmark}{\ding{55}} % ✗
\usepackage{xcolor}
\usepackage[most]{tcolorbox}
\usepackage{enumitem}
\tcbuselibrary{breakable}

\definecolor{darkgreen}{RGB}{0,200,0}
\definecolor{purple}{RGB}{153, 51, 204}
\definecolor{blue}{RGB}{37, 99, 235}
\definecolor{lightaccentblue}{RGB}{219, 234, 254}
\definecolor{lightaccentgreen}{HTML}{64C9D2}

\definecolor{azure}{rgb}{0.94, 0.97, 1.0} 
\definecolor{lavender}{rgb}{0.96, 0.94, 1.0}

\newcommand{\tablestyle}[2]{\setlength{\tabcolsep}{#1}\renewcommand{\arraystretch}{#2}\centering\footnotesize}
\usepackage{hyperref}
\usepackage{multirow}

\usepackage[preprint]{icml2026}

\usepackage{amsmath}
\usepackage{amssymb}
\usepackage{mathtools}
\usepackage{amsthm}

\usepackage[capitalize,noabbrev]{cleveref}

\theoremstyle{plain}

\theoremstyle{definition}

\theoremstyle{remark}

\usepackage[textsize=tiny]{todonotes}

\icmltitlerunning{ViGoR-Bench: How Far Are Visual Generative Models From Zero-Shot Visual Reasoners?}

\begin{document}

\twocolumn[
  \icmltitle{ViGoR-Bench: How Far Are Visual Generative Models
From Zero-Shot\\ Visual Reasoners?}

  % It is OKAY to include author information, even for blind submissions: the
  % style file will automatically remove it for you unless you've provided
  % the [accepted] option to the icml2026 package.

  % List of affiliations: The first argument should be a (short) identifier you
  % will use later to specify author affiliations Academic affiliations
  % should list Department, University, City, Region, Country Industry
  % affiliations should list Company, City, Region, Country

  % You can specify symbols, otherwise they are numbered in order. Ideally, you
  % should not use this facility. Affiliations will be numbered in order of
  % appearance and this is the preferred way.
  \icmlsetsymbol{equal}{*}
  \icmlsetsymbol{projectlead}{\ddag}
  \icmlsetsymbol{correspond}{\dag}

  \begin{icmlauthorlist}
    \icmlauthor{Haonan Han}{tsinghua,meituan}
    \icmlauthor{Jiancheng Huang}{meituan}
    \icmlauthor{Xiaopeng Sun}{meituan}
    \icmlauthor{Junyan He}{meituan,projectlead}
    \icmlauthor{Rui Yang}{HKU}
    \icmlauthor{Jie Hu}{meituan}\\
    \icmlauthor{Xiaojiang Peng}{SIAT}
    \icmlauthor{Lin Ma}{meituan}
    \icmlauthor{Xiaoming Wei}{meituan}
    \icmlauthor{Xiu Li}{tsinghua,correspond}
  \end{icmlauthorlist}

  \icmlaffiliation{tsinghua}{Tsinghua University}
  \icmlaffiliation{meituan}{Meituan M17}
  \icmlaffiliation{HKU}{The University of Hong Kong}
  \icmlaffiliation{SIAT}{Shenzhen Institutes of Advanced Technology, Chinese Academy of Sciences}
    
  \icmlcorrespondingauthor{Xiu Li}{li.xiu@sz.tsinghua.edu.cn}

  % You may provide any keywords that you find helpful for describing your
  % paper; these are used to populate the "keywords" metadata in the PDF but
  % will not be shown in the document
  \icmlkeywords{Machine Learning, ICML}

  \vskip 0.3in
]
% this must go after the closing bracket ] following \twocolumn[ ...

% This command actually creates the footnote in the first column listing the
% affiliations and the copyright notice. The command takes one argument, which
% is text to display at the start of the footnote. The \icmlEqualContribution
% command is standard text for equal contribution. Remove it (just {}) if you
% do not need this facility.

% Use ONE of the following lines. DO NOT remove the command.
% If you have no special notice, KEEP empty braces:
\printAffiliationsAndNotice{}  % no special notice (required even if empty)
% Or, if applicable, use the standard equal contribution text:
% \printAffiliationsAndNotice{\icmlEqualContribution}

\begin{abstract}
%While modern AIGC models achieve stunning visual fidelity, they often mask a ``logical desert,'' failing tasks that require physical, causal, or complex spatial reasoning. 
Beneath the stunning visual fidelity of modern AIGC models lies a ``logical desert", where systems fail tasks that require physical, causal, or complex spatial reasoning. Current evaluations largely rely on superficial metrics or fragmented benchmarks, creating a ``performance mirage'' that overlooks the generative process. To address this, we introduce \textbf{ViGoR} (\textbf{Vi}sion-\textbf{G}enerative \textbf{R}easoning-centric Benchmark), a unified framework designed to dismantle this mirage. ViGoR distinguishes itself through four key innovations: 1) \textbf{holistic cross-modal coverage} bridging Image-to-Image and Video tasks; 2) a \textbf{dual-track mechanism} evaluating both intermediate processes and final results; 3) an \textbf{evidence-grounded automated judge} ensuring high human alignment; and 4) \textbf{granular diagnostic analysis} that decomposes performance into fine-grained cognitive dimensions. Experiments on over 20 leading models reveal that even state-of-the-art systems harbor significant reasoning deficits, establishing ViGoR as a critical ``stress test'' for the next generation of intelligent vision models. The demo have been available at \url{https://vincenthancoder.github.io/ViGoR-Bench/}.
\end{abstract}
\section{Introduction}

\begin{figure*}[t!]
    \centering 
    \includegraphics[width=0.95\textwidth]{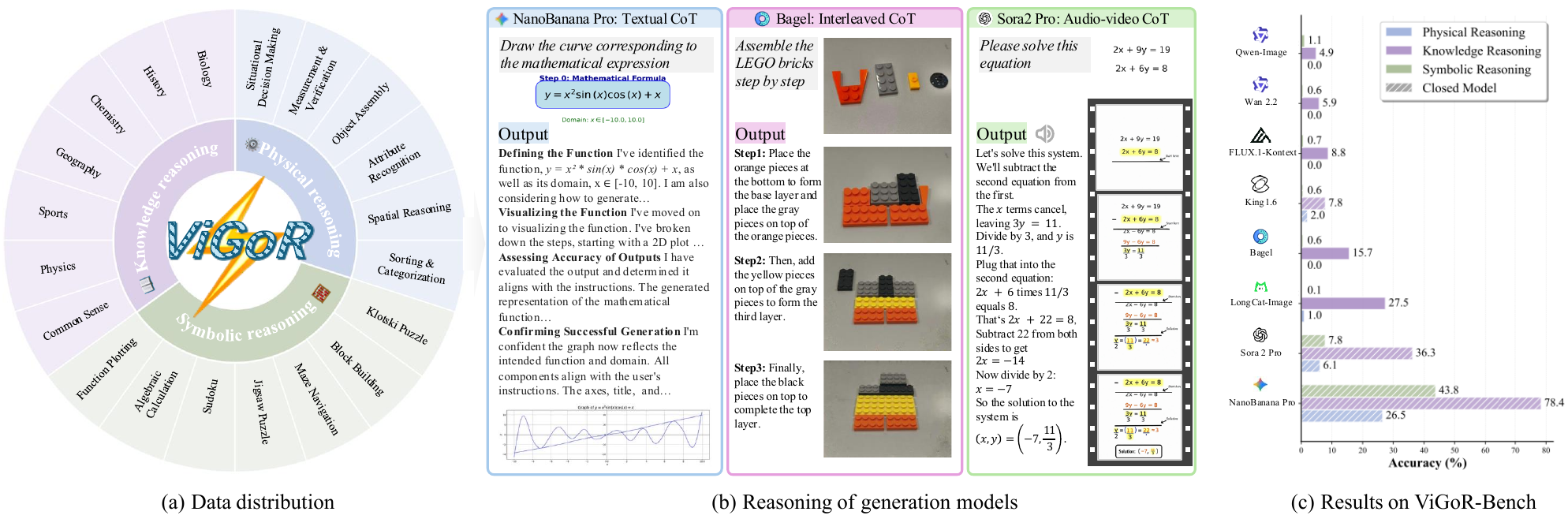}
    \caption{\textbf{An overview of ViGoR-Bench.} (a) The data distribution across various domains. (b) Examples of the reasoning process from generation models. (c) Performance comparison of leading models on ViGoR-Bench.}
    \label{fig:teaser}
\vspace{-0.15in}
\end{figure*}

%The landscape of AI Generated Content (AIGC) has undergone a paradigm shift, moving from simple pixel-level work to advanced content generation. This progress reflects an architectural leap from early GANs to modern, high-capacity generative systems~\cite{}. State-of-the-art models—ranging from Flux and DALL-E 3 to systems like SORA2—have reached impressive visual fidelity, rendering them nearly indistinguishable from human-authored content. However, this high visual quality often hides a core deficit: a ``logical desert." Despite their realism, these models still struggle with tasks that require world knowledge, physical logic, or complex, multi-step reasoning.
Artificial Intelligence-Generated Content (AIGC) has witnessed remarkable growth, evolving from basic pixel-level synthesis to the production of highly sophisticated content. This progress reflects an architectural leap from early GANs~\cite{gan} to modern, high-capacity generative systems~\cite{rombach2021highresolution, blattmann2023stable,flux-2-2025,cao2025hunyuanimage,seedream2025seedream}. 
%Current state-of-the-art models—ranging from Flux and DALL-E 3 to systems like Sora—have achieved impressive visual fidelity, rendering them nearly indistinguishable from human-authored content. 
%However, such visual excellence often masks a core deficit: a ``logical desert". Despite their outward realism, these models still struggle with tasks that require deep world knowledge, physical logic, or complex, multi-step reasoning. The traditional reliance on quantitative metrics like CLIP-Score and FID has fostered a performance mirage: specifically, CLIP-Score acts as a "bag-of-words" matcher blind to spatial logic, while FID prioritizes statistical distribution over structural integrity. 
However, such visual excellence often masks a ``logical desert": beneath a facade of photorealism, models crumble when faced with tasks requiring deep physical laws or causal reasoning. This deficit is obscured by a performance mirage fostered by traditional metrics like CLIP-Score~\cite{hessel2021clipscore} and FID~\cite{heusel2017gans}, which prioritize semantic alignment and statistical fidelity over true structural integrity. 
Notably, a generated image can achieve high statistical similarity to real data while still harboring absurd physical glitches. Consequently, existing metrics fail to distinguish between a model that truly ``understands" the physical world and one that merely performs high-dimensional probability tiling. This underscores an urgent need to pivot from evaluating fidelity to rigorously assessing the generative reasoning capabilities that define true visual intelligence.
In response to this ``logical desert", benchmarks have evolved beyond pixel fidelity to probe cognitive depth along three dimensions: knowledge breadth, where GenExam~\cite{wang2025genexam} audits factual consistency; physical causality, exemplified by KRIS-Bench's~\cite{wu2025kris} focus on dynamic editing; and process-oriented temporal logic, where RULER-Bench~\cite{he2025ruler} and MME-CoF~\cite{guo2025video} validate the continuous ``reasoning chain" in dynamic tasks.
Despite these advances, the current evaluation landscape remains fragmented. As illustrated in Table~\ref{tab:benchmark_overview}, existing benchmarks typically operate in \textit{silos-narrowly} restricted to either Image-to-Image (I2I) editing or Video generation (I2V)---lacking a unified mechanism to assess reasoning across diverse modalities. Critically, most frameworks fail to distinguish between the final outcome (the ``what'') and the generative process (the ``how''). This conceptual gap is mirrored in the evaluation domain, while the ``VLM-as-a-Judge'' paradigm (e.g., GPT-4o~\cite{openai2025gpt4o}, Gemini 2.5 Pro~\cite{google2025gemini}) has emerged as the \textit{de facto} standard for scalable evaluation, achieving robust human alignment across multifaceted reasoning tasks remains a persistent bottleneck. \textbf{To address these limitations, we introduce \textbf{ViGoR-Bench} (\textbf{Vi}sion-\textbf{G}enerative \textbf{R}easoning-centric Benchmark),} a comprehensive evaluation framework designed to unveil the ``performance mirage.'' Unlike its predecessors, ViGoR-Bench is defined by three key innovations:

\begin{itemize}[leftmargin=*, noitemsep, topsep=2pt]
     \item \textbf{Holistic Cross-Modal Coverage:} ViGoR-Bench is the first benchmark to bridge the divide between I2I, Sequential I2I (I2Is), and I2V tasks. As shown in Figure~\ref{fig:teaser}~(a), it encompasses 20 distinct dimensions---ranging from physical mechanics and social commonsense to complex spatial planning---providing a unified testbed for generative intelligence.
    
    \noindent  \item \textbf{Dual-Track Process-Outcome Evaluation:} Moving beyond binary outcome assessment, ViGoR-Bench implements a rigorous \textit{Process plus Result} scoring pipeline, which evaluates not only the final output but also verifies whether intermediate states and logic adhere to physical laws and causal consistency.
    
    \item \textbf{Evidence-Grounded Automated Alignment:} Leveraging a multi-agent ``Evidence-Grounded'' judge system, ViGoR-Bench mitigates the subjectivity of LLM evaluators and achieves unprecedented alignment with human experts, demonstrating superior MAE (Mean Absolute Error) and Pearson correlations.%, ensuring granular and objective analysis.
    
    \item \textbf{Granular Diagnostic Analysis:} Moving beyond standard leaderboards, ViGoR introduces a structured analysis of reasoning failures. It breaks down model performance into specific cognitive dimensions, enabling researchers to pinpoint specific reasoning gaps rather than relying on a single aggregate score.

    % 我们的benchmark里面的数据貌似反常识逻辑的数据有应该没有专门构造，不过Gemini提示这个点是挺有意思的，也是对生成模型的一个攻击
    %\item \textbf{Adversarial Logic Scenarios:} To counteract model reliance on rote memorization, ViGoR incorporates ``counter-intuitive'' and adversarial scenarios designed to break statistical correlations. This ensures that high scores reflect genuine reasoning capabilities rather than the retrieval of common patterns from the training distribution.
\end{itemize}

Through extensive experiments on over 20 leading generative models, we demonstrate that ViGoR-Bench serves as a critical stress test for visual reasoning. Our results reveal that even models with superior visual fidelity can harbor significant deficits in world-knowledge reasoning. This underscores the need for a paradigm shift toward more intelligent vision foundation models, as illustrated in Figure~\ref{fig:teaser}.
Our key findings are as follows:
\begin{itemize}[leftmargin=*, noitemsep, topsep=2pt]
    \item Proprietary models maintain a significant performance lead over their open-source counterparts (see Figure~\ref{fig:teaser}).
    \item Explicit Chain-of-Thought (CoT) prompting enhances the interpretability of the generation process but it does not guarantee an improvement in final accuracy.
    \item Video generation models often exhibit an ``Illusion of Reasoning'', where apparent logical consistency does not hold up to rigorous evaluation.
\end{itemize}
Furthermore, by training models on a specific subtask, we observed that:
\begin{itemize}[leftmargin=*, noitemsep, topsep=2pt]
    \item Reward-driven Reinforcement Learning (RL) demonstrates superior potential in advancing visual reasoning capabilities where Supervised Fine-Tuning (SFT) exhibits saturation.
    \item Training on more challenging, Out-of-Distribution (OOD) data enhances a model's generalization performance on simpler, in-distribution visual reasoning tasks.
\end{itemize}

\begin{table}[t!]
\centering
\caption{
Overview of benchmark properties and evaluation settings. Veo$^{\dagger}$~\cite{wiedemer2025videomodelszeroshotlearners} conducted quantitative evaluations across seven distinct task categories. 
}
\label{tab:benchmark_overview}
\tablestyle{5.0pt}{1.0}
\resizebox{\columnwidth}{!}{
\begin{tabular}{l|c|c c| c c c| c c}
\toprule
\textbf{Benchmark}
& \textbf{Tasks}
& \multicolumn{2}{c|}{\textbf{Reference}}
& \multicolumn{3}{c|}{\textbf{Type}}
& \multicolumn{2}{c}{\textbf{Evaluation}} \\
\cmidrule(lr){3-4}
\cmidrule(lr){5-7}
\cmidrule(lr){8-9}
& & 
\textit{\small{Img}}
& \textit{\small{Text}}
& \textit{\small{I2I}}
& \textit{\small{I2Is$^{\star}$}}
& \textit{\small{I2V}}
& \textit{\small{Process}}
& \textit{\small{Result}} \\
\midrule

% ===== rows go here =====
RISE        & -- & \textcolor{red}{\xmark} & \textcolor{darkgreen}{\cmark} & \textcolor{darkgreen}{\cmark} & \textcolor{red}{\xmark} & \textcolor{red}{\xmark} & \textcolor{red}{\xmark} & \textcolor{darkgreen}{\cmark} \\
KRIS    & 22  & \textcolor{red}{\xmark} & \textcolor{darkgreen}{\cmark} & \textcolor{darkgreen}{\cmark} & \textcolor{red}{\xmark} & \textcolor{red}{\xmark} & \textcolor{red}{\xmark} & \textcolor{darkgreen}{\cmark} \\
GIR-Edit & 3 & \textcolor{darkgreen}{\cmark} & \textcolor{darkgreen}{\cmark} & \textcolor{darkgreen}{\cmark} & \textcolor{red}{\xmark} & \textcolor{red}{\xmark} & \textcolor{red}{\xmark} & \textcolor{darkgreen}{\cmark} \\
UniREdit & 18 & \textcolor{darkgreen}{\cmark} & \textcolor{darkgreen}{\cmark} & \textcolor{darkgreen}{\cmark} & \textcolor{darkgreen}{\cmark} & \textcolor{red}{\xmark} & \textcolor{darkgreen}{\cmark} & \textcolor{darkgreen}{\cmark} \\
WiseEdit & -- & \textcolor{red}{\xmark} & \textcolor{darkgreen}{\cmark} & \textcolor{darkgreen}{\cmark} & \textcolor{darkgreen}{\cmark} & \textcolor{red}{\xmark} & \textcolor{red}{\xmark} & \textcolor{darkgreen}{\cmark} \\
\midrule
Veo$^{\dagger}$ & 7$^{\ddagger}$ & \textcolor{darkgreen}{\cmark} & \textcolor{red}{\xmark} & \textcolor{red}{\xmark} & \textcolor{red}{\xmark} & \textcolor{darkgreen}{\cmark} & \textcolor{darkgreen}{\cmark} & \textcolor{darkgreen}{\cmark} \\
MME-CoF & 12 & \textcolor{red}{\xmark} & \textcolor{darkgreen}{\cmark} & \textcolor{red}{\xmark} & \textcolor{red}{\xmark} & \textcolor{darkgreen}{\cmark} & \textcolor{darkgreen}{\cmark} & \textcolor{darkgreen}{\cmark} \\
RULER & 40 & \textcolor{darkgreen}{\cmark} & \textcolor{darkgreen}{\cmark} & \textcolor{red}{\xmark} & \textcolor{red}{\xmark} & \textcolor{darkgreen}{\cmark} & \textcolor{darkgreen}{\cmark} & \textcolor{red}{\xmark} \\
% ===== rows go here =====

\midrule
\rowcolor{gray!25}
\textbf{ViGoR (ours)} 
& \textbf{20}
& \textcolor{darkgreen}{\cmark} & \textcolor{darkgreen}{\cmark}
& \textcolor{darkgreen}{\cmark} & \textcolor{darkgreen}{\cmark} & \textcolor{darkgreen}{\cmark}
& \textcolor{darkgreen}{\cmark} & \textcolor{darkgreen}{\cmark} \\
\bottomrule
\end{tabular}
}
\end{table}

\section{Related Work}
\subsection{Visual Generative Model}

\textit{Text-to-Image and Editing Models.} T2I generation has laid the bedrock for visual synthesis, driven by foundational architectures like Stable Diffusion~\cite{rombach2021highresolution} and Flux~\cite{flux-2-2025} which revolutionized generation efficiency. Building on this, conditional editing has evolved through three paradigms: (1) \textbf{Spatial Control}, where ControlNet~\cite{zhang2023adding} and T2I-Adapter~\cite{mou2024t2i} inject structural guidance via additional conditions; (2) \textbf{Attention Manipulation}, exemplified by Prompt-to-Prompt~\cite{hertz2022prompt}, which modifies cross-attention maps for zero-shot editing; and (3) \textbf{Instruction Following}. Recently, the latter has seen rapid advancements with models like Qwen-Image-Edit~\cite{wu2025qwenimagetechnicalreport} and Z-Image~\cite{team2025zimage} optimizing semantic precision, while Seed-Edit~\cite{wang2025seededit} and the nano-banana family (Gemini 2.5)~\cite{google2025nanobanana} push the boundaries of sequential consistency and character fidelity.

% \textit{Unified Vision Models.} A recent trend aims to unify understanding and generation 
% within single architectures. Unified-IO [Lu et al., 2022] and Unified-IO 2 [Lu et al., 2023] 
% handle diverse vision tasks through a sequence-to-sequence formulation. Chameleon [Team, 2024] 
% proposes an early-fusion token-based mixed-modal architecture supporting both 
% understanding and generation. Emu [Sun et al., 2023] and Emu2 [Sun et al., 2024] 
% train generalist multimodal models with unified objectives. SEED [Ge et al., 2023] 
% and SEED-X [Ge et al., 2024] introduce a unified visual tokenizer for both discrete 
% and continuous representations. Transfusion [Zhou et al., 2024] combines language 
% modeling with diffusion for multimodal generation. Show-o [Xie et al., 2024] 
% proposes a unified transformer for multimodal understanding and generation with 
% next-token prediction, enabling flexible control through natural language.  These unified models \textbf{provide 
% an ideal testbed for comparing reasoning capabilities across understanding and 
% generation tasks}, which our benchmark specifically targets.

\textit{Unified Vision Models.} A recent trend unifies understanding and generation 
within single architectures, enabling reasoning across modalities. Early works like 
Unified-IO~\citep{Lu2023UnifiedIO, Lu2024UnifiedIO2} handle diverse vision tasks through sequence-to-sequence formulation. Interleaved image-text models represent a paradigm shift: Chameleon~\citep{Lu2023Chameleon} employs early-fusion token-based architecture for mixed-modal processing; Emu series~\cite{Sun2024Emu1, Sun2024Emu2} and SEED-X~\cite{Ge2024SEEDX} demonstrate unified visual tokenization for coherent multimodal reasoning; Show-o~\cite{xie2024showo} achieves flexible control through next-token prediction across modalities. Recent autoregressive approaches like Janus~\cite{wu2024janus}, Anole~\cite{chern2024anole}, and VILA-U~\cite{Wu2025VILA-U} decouple or unify visual pathways for enhanced reasoning transfer. BAGEL~\cite{deng2025bagel} demonstrates that scaling unified models with large-scale interleaved data reveals emergent reasoning capabilities, including world-modeling tasks like multiview synthesis and visual navigation. Hybrid architectures such as Transfusion~\cite{Zhou2025Transfusion}, Meissonic~\cite{bai2024meissonic}, and Lumina-mGPT~\cite{liu2024lumina-mgpt} combine different modeling paradigms for flexible any-to-any generation. %**These models exhibit qualitatively  different capabilities**: they can analyze images, reason about changes, and generate corresponding outputs—effectively "thinking with images" through alternating understanding  and synthesis, going beyond traditional single-direction generation or understanding alone.

\textit{Video Generation Models.} Video generation extends to temporal sequences, demanding reasoning about dynamics, physics, and causality. Building upon early diffusion-based work [Singer et al., 2023; Ho et al., 2022], recent models including Sora series \citep{openai2024sora, openai2025sora2}, Veo3~\cite{googledeepmind2024veo3}, Kling~\cite{kuaishou2024kling},  VideoPoet~\cite{Kondratyuk2024VideoPoet}, seedance~\cite{bytedance2025seedance,chen2025seedance}, Lumiere~\citep{bar2024lumiere} , and Stable Diffusion 4.0~\cite{yao2025sv4d} demonstrate remarkable temporal coherence and physical plausibility. %\textbf{These advances suggest sophisticated implicit reasoning about world dynamics, object permanence, and causal relationships}, providing ideal candidates for evaluating temporal and physical reasoning capabilities.

\subsection{Benchmarks for Visual Generation}
\textit{Perceptual Quality of Generation.} Current benchmarks for generative models predominantly focus on generation quality 
(FID~\cite{heusel2017gans}, IS~\cite{salimans2016improved}, CLIP Score~\cite{hessel2021clipscore}) 
or specific compositional aspects (T2I-CompBench~\cite{huang2023t2i}, GenEval~\cite{ghosh2023geneval}, 
TIFA~\cite{hu2023tifa}).

\textit{Reasoning-Centric Evaluation.} Generative model evaluation is shifting from perceptual quality to cognitive reasoning. Image benchmarks like GenExam~\cite{wang2025genexam}, SridBench~\cite{chang2025SridBench}, and WISE now integrate multidisciplinary knowledge, while GIR-BENCH~\cite{Li2025GIRBench} and KRIS-Bench~\cite{wu2025kris} demand logical consistency. In the video domain, MME-COF~\cite{guo2025video} and Reasoning via Video test temporal logic, while PICABench~\cite{pu2025picabench} and RULER-Bench~\cite{he2025ruler} rigorously evaluate adherence to physical laws. Methodologically, OneIG-Bench~\cite{chang2025oneig} and WiseEdit~\cite{pan2025wiseedit} establish VLM-as-a-Judge as the standard paradigm. To ensure objectivity, works like RISEBench~\cite{zhao2025envisioning} and UniREditBench~\cite{han2025unireditbench} validate Human-LLM alignment. %Metrics have evolved beyond aesthetics to prioritize Instruction Following and Visual Consistency. Collectively, research such as PicWorld and MotionEdit aims to quantify the "World Model" capabilities—specifically physics, logic, and causal reasoning—within generative AI.
\section{ViGoR-Bench}
\begin{figure*}[h]
    \centering 
    \includegraphics[width=0.95\textwidth]{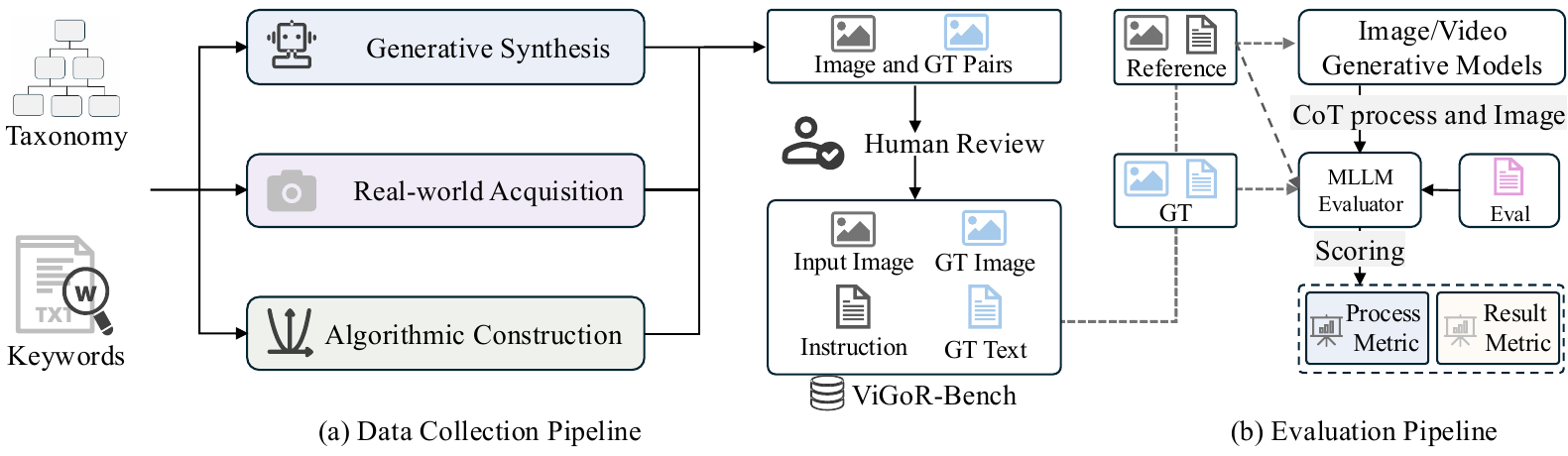}
    \caption{\textbf{An overview of the ViGoR-Bench construction and evaluation pipelines.} (a) The benchmark dataset is constructed through a three-pronged approach: generative synthesis, real-world acquisition, and algorithmic generation. All data undergoes human review to establish definitive image-ground truth (GT) pairs. (b) For evaluation, a Multimodal Large Language Model (MLLM) is employed as an automated judge. Conditioned on the ground-truth image, the MLLM assesses both the Chain-of-Thought (CoT) reasoning process and the final output of generative models for images and videos.}
    \label{fig:pipeline}
    % \vspace{-0.2in}
\end{figure*}

To comprehensively evaluate the reasoning capabilities of generative models, we constructed a diverse benchmark comprising three primary domains: Physical Reasoning, Knowledge Reasoning, and Symbolic Reasoning. Figure~\ref{fig:pipeline}~(a) summarizes the taxonomy of our benchmark.

\subsection{Data Engine}

\begin{figure*}[t]
    \
    \centering
    \includegraphics[width=0.95\textwidth]{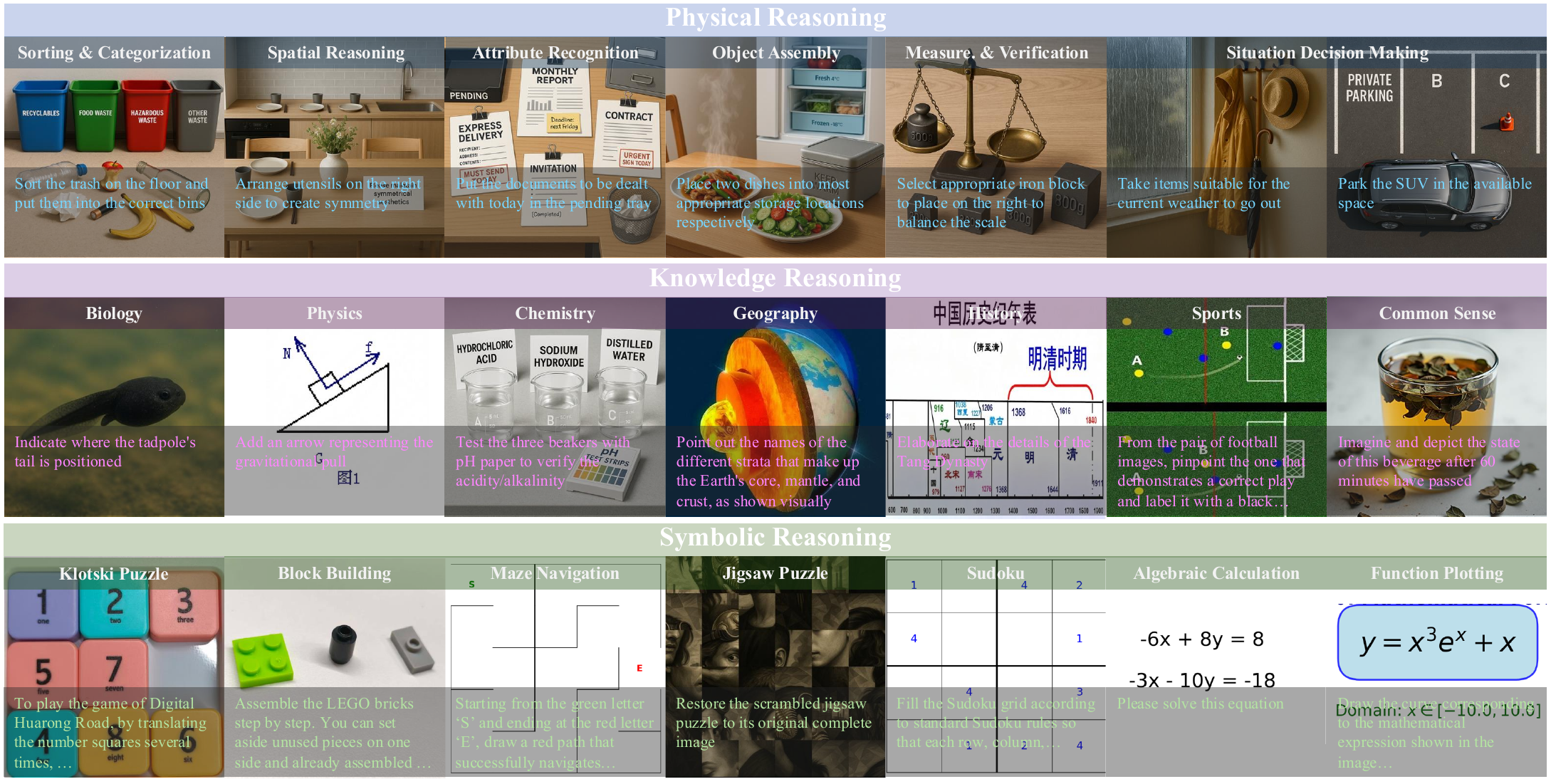}
    \caption{\textbf{Overview of the ViGoR-Bench task suite.} We present representative demo cases and their corresponding editing instructions across 20 distinct sub-tasks. These tasks are hierarchically organized into three primary reasoning domains: Physical Reasoning, Knowledge Reasoning, and Symbolic Reasoning.}
    \label{fig:demo}
% \vspace{-0.2in}
\end{figure*}

To build the ViGoR-Bench, we designed a data construction pipeline tailored to the unique characteristics of each reasoning domain. Our approach integrates three distinct construction paradigms: (1) Generative Synthesis, which leverages large language models and image generation models to create high-fidelity physical scenarios; (2) Real-world Acquisition, involving authoritative web curation and manual photography to ensure alignment with reality; and (3) Algorithmic Construction, utilizing rule-based engines to produce logically rigorous samples. Figure~\ref{fig:demo} presents representative examples curated through our data collection pipeline, spanning 20 distinct subdomains across three reasoning domains. Crucially, to guarantee the correctness and reliability of the benchmark, we incorporate a rigorous post-processing verification stage. This includes human-in-the-loop review for semantic consistency and symbolic solver validation for mathematical precision. Unlike previous benchmarks for generative model evaluation, our benchmark provides both referenced ground-truth images and human-verified ground-truth captions where applicable. Below, we detail the sample collection strategies for each subdomain.

\noindent \textbf{Physical Reasoning.}
The Physical Reasoning subset encompasses scenarios requiring embodied intelligence, including tasks such as Sorting, Categorization, Spatial Reasoning, Attribute Recognition, Object Assembly, Measurement $\&$ Verification, and Situational Decision Making. 
Due to the high cost and complexity of acquiring diverse real-world embodied data, we employ a generative pipeline. First, detailed textual descriptions of physical scenarios and tasks are composed, then further enriched using large language models. These descriptions serve as prompts for state-of-the-art generative models, i.e., NanoBanana-Pro~\cite{google2025nanobananapro}, which synthesize high-fidelity input images. All generated images are subsequently verified by human annotators for plausibility and relevance.
Since all input images are generated, no corresponding ground-truth images exist. Instead, we provide textual ground-truth answers, which are manually annotated or verified by human experts to ensure logical consistency with the visual input.

\noindent \textbf{Knowledge Reasoning.}
This domain assesses the ability to reason over world knowledge, spanning disciplines such as Biology, Physics, Chemistry, Geography, History, Sports, and Common Sense.
A substantial portion of the data is curated from authoritative educational websites and scientific repositories to ensure factual accuracy. 
All samples are accompanied by human-verified textual answers. For cases where paired datasets exist (e.g., ``before-and-after" scientific phenomena), the original ground-truth images are preserved. For other samples, only textual ground-truth is provided.

\noindent \textbf{Symbolic Reasoning.}
These tasks require precise logical manipulation. For different types of data, we use different approaches.
For Physical Puzzles (Klotski Puzzle, Block Building), we emphasize visual realism to evaluate the model's alignment between perception and reasoning. Data is collected in physical environments, where annotators manually solve puzzles and capture the solved states as ground-truth images.
For abstract logic tasks (e.g., Sudoku, Maze Navigation, Jigsaw Puzzle, Function Plotting), we employ rule-based algorithms to ensure mathematical rigor and uniqueness of solutions. Input and ground-truth images are algorithmically generated, with no textual ground-truth provided.
For algebraic calculation tasks, we write equations, covering linear and quadratic forms, using large language models, and validate their solutions via symbolic solvers. Both the equations and their solutions are rendered as images to serve as input and ground-truth, respectively.
Similarly, for function plotting tasks, two-dimensional function expressions are generated. The corresponding curves are plotted using Matplotlib to produce ground-truth images.

\subsection{Evaluation Protocol}
\label{sec:evaluation_protocol}

To provide a comprehensive assessment of visual reasoning capabilities, we establish a dual-track evaluation protocol comprising \textbf{Process Metrics} and \textbf{Result Metrics}, as shown in Figure~\ref{fig:pipeline}~(b). This design allows us to scrutinize not only the correctness of the final solution but also the logical coherence of the intermediate reasoning steps. We employ Gemini-2.5-Pro~\cite{gemini25pushingfrontier} as our \textbf{VLM-as-a-Judge} evaluator due to its advanced multimodal understanding capabilities.

\textbf{Process Metric.} This set of metrics is designed to evaluate dynamic outputs, such as video generation models or ``thinking" models that produce intermediate reasoning frames. It assesses the quality and trajectory of the reasoning process. The evaluation is formulated as a function of the input and the generated sequence:
\begin{equation}
    \mathbf{S}_{\text{Process}} = \text{VLM}(I, P, O_{\text{seq}}, R_i, R_t, \mathcal{T}_{\text{Process}})
\end{equation}
where $I$ denotes the input image, $P$ represents the editing prompt, and $O_{\text{seq}}$ is the model's output sequence (intermediate frames or video). $R_i$ and $R_t$ denote the visual Ground Truth (GT) image and the textual GT reference, respectively. $\mathcal{T}_{\text{Process}}$ represents the specific evaluation template containing scoring criteria. The output score vector consists of four dimensions:
\begin{equation}
    \mathbf{S}_{\text{Process}} = [S_{\text{BC}}, S_{\text{RO}}, S_{\text{VQ}}, S_{\text{RA}}]
\end{equation}
The final aggregated score is calculated as the mean of these components:
\begin{equation}
    S_{\text{Avg}}^{\text{Process}} = \frac{1}{4} (S_{\text{BC}} + S_{\text{RO}} + S_{\text{VQ}} + S_{\text{RA}})
\end{equation}
Each sub-metric is scored on a continuous scale from 0 to 100, defined as follows:
\begin{itemize}[leftmargin=*, noitemsep, topsep=2pt]
    \item \textbf{Background Consistency ($S_{\text{BC}}$):} Measures the extent to which the main structure of the input image $I$ is preserved across the output sequence $O_{\text{seq}}$. It penalizes unintended modifications to regions unrelated to the reasoning task.
    \item \textbf{Rule Obey ($S_{\text{RO}}$):} Assesses the percentage of frames where edits strictly adhere to the constraints and requirements specified in the instruction $P$.
    \item \textbf{Visual Quality ($S_{\text{VQ}}$):} Evaluates the fidelity of the generated frames, checking for clarity, sharpness, and the absence of temporal flickering, noise, or artifacts.
    \item \textbf{Reasoning Accuracy ($S_{\text{RA}}$):} Evaluates the efficacy of the progressive edits. It determines whether the model's modifications effectively progress toward the correct solution as defined by $R_i$ and $R_t$ (analogous to ``Beneficial Action").
\end{itemize}

%>>>>>>>>>>>>>>>>>>>>>>>>reliability exp>>>>>>>>>>>>>>>>>>>>>>>>

\begin{table}[t]
\centering
\caption{\textbf{Reliability Analysis of VLM-as-a-Judge.} We evaluate the alignment between Gemini-2.5-Pro~\cite{gemini25pushingfrontier} and human experts on a random subset. The table compares performance with and without Ground Truth references across both Process and Result metrics.}
\label{tab:evaluation-results}
\tablestyle{5.0pt}{1.0}
\resizebox{\columnwidth}{!}{%
    \setlength{\tabcolsep}{5pt}
    \begin{tabular}{l|l|c|ccc}
    \toprule
    Type & Evaluator & GT Ref. & MAE $\downarrow$ & Acc $\uparrow$ & Var $\downarrow$ \\
    \midrule
    \multirow{3}{*}{Process} 
     & \small Human & \textcolor{darkgreen}{\cmark} & \small{0.000} & \small{1.000} & \small{0.051} \\
     & \small Gemini-2.5-Pro & \textcolor{red}{\xmark} & \small{0.319} & \small{0.680} & \small{0.039} \\
     & \cellcolor{gray!25}\small Gemini-2.5-Pro & \cellcolor{gray!25}\textcolor{darkgreen}{\cmark} & \cellcolor{gray!25}\small{\textbf{0.267}} & \cellcolor{gray!25}\small{\textbf{0.733}} & \cellcolor{gray!25}\small{\textbf{0.034}} \\ 
    \midrule 
    \addlinespace[0.5ex] 
    \multirow{3}{*}{Result} 
     & \small Human & \textcolor{darkgreen}{\cmark} & \small{0.000} & \small{1.000} & \small{0.011} \\
     & \small Gemini-2.5-Pro & \textcolor{red}{\xmark} & \small{0.294} & \small{0.705} & \small{0.034} \\
     & \cellcolor{gray!25}\small Gemini-2.5-Pro & \cellcolor{gray!25}\textcolor{darkgreen}{\cmark} & \cellcolor{gray!25}\small{\textbf{0.213}} & \cellcolor{gray!25}\small{\textbf{0.786}} & \cellcolor{gray!25}\small{\textbf{0.029}} \\
    \bottomrule
    \end{tabular}%
}
\end{table}

%>>>>>>>>>>>>>>>>>>>>>>>>reliability exp>>>>>>>>>>>>>>>>>>>>>>>>

\textbf{Result Metric.} This set of metrics targets the final output—either the static image generated by image editing models or the final frame of a reasoning sequence. It focuses on the validity of the final solution. Similar to the process evaluation, the scoring function is defined as:
\begin{equation}
    \mathbf{S}_{\text{Result}} = \text{VLM}(I, P, O_{\text{final}}, R_i, R_t, \mathcal{T}_{\text{Result}})
\end{equation}
Here, $O_{\text{final}}$ represents the final generated image. The resulting score vector is composed of:
\begin{equation}
    \mathbf{S}_{\text{Result}} = [S_{\text{BC}}, S_{\text{RO}}, S_{\text{VQ}}, S_{\text{RS}}]
\end{equation}
The average performance is derived as:
\begin{equation}
    S_{\text{Avg}}^{\text{Result}} = \frac{1}{4} (S_{\text{BC}} + S_{\text{RO}} + S_{\text{VQ}} + S_{\text{RS}})
\end{equation}
Distinct from the process metrics, the Result Metrics employ a binary scoring system $\{0, 1\}$ to provide a rigorous pass/fail assessment:
\begin{itemize}[leftmargin=*, noitemsep, topsep=2pt]
    \item \textbf{Background Consistency ($S_{\text{BC}}$):} A binary check on whether the output image $O_{\text{final}}$ retains the structural integrity of the input $I$, ensuring irrelevant areas remain untouched.
    \item \textbf{Rule Obey ($S_{\text{RO}}$):} Determines if the result complies with the explicit instructions given in $P$ while adhering to essential reasoning constraints (e.g., avoiding wall penetration in maze navigation).
    \item \textbf{Visual Quality ($S_{\text{VQ}}$):} Verifies if the final output maintains high realism and is free from degradation, distortions, or physical implausibility.
    \item \textbf{Reasoning Success ($S_{\text{RS}}$):} The critical measure of task completion. It evaluates whether the final state matches the reference answer ($R_t$) or the reference image ($R_i$), signifying a correct solution to the reasoning problem.
\end{itemize}

\subsection{Reliability Analysis}
\label{sec:reliability}

To validate the trustworthiness of our evaluation pipeline, we conducted a rigorous meta-evaluation comparing our VLM-as-a-Judge against human experts. 

%>>>>>>>>>>>>>>>>>>>>>>>>>Main Result>>>>>>>>>>>>>>>>>>>>>>>>
\setlength{\aboverulesep}{0pt}
\setlength{\belowrulesep}{0pt}
\begin{table*}[h]
\vspace{-0.1in}
\centering
\tablestyle{5.0pt}{1.0}
\caption{\textbf{Main experimental results across different metrics.}
Process metrics evaluate generative process reasoning quality, while final result metrics assess the final output; all values are reported in percentage scale.
\textbf{OS}$^{\dagger}$ indicates \textbf{open-sourced} status.
Within each model category, the best result for each metric is marked in \textbf{bold}.
\textcolor{purple}{\textbf{Bold values}} denote the overall best performance across all results.}
\label{tab:main_results}
\resizebox{\textwidth}{!}{
\begin{tabular}{
c|c|c|c |
c c c >{\columncolor{lightaccentblue}}c >{\columncolor{lightaccentgreen!20}}c |
c c c >{\columncolor{lightaccentblue}}c >{\columncolor{lightaccentgreen!20}}c
}
\toprule
\textbf{Type} & \textbf{Process} & \textbf{Model} & \textbf{OS$^{\dagger}$}
& \multicolumn{5}{c|}{\textbf{Process Metric} (\%)}
& \multicolumn{5}{c}{\textbf{Result Metric} (\%)} \\
\cmidrule(lr){5-9} \cmidrule(lr){10-14}
& & & 
& \textit{\small{BC$\uparrow$}} & \textit{\small{RO$\uparrow$}} & \textit{\small{VQ$\uparrow$}} & \textit{\small{\textbf{RA}$\uparrow$}} & \small{\textbf{Avg}}
& \textit{\small{BC$\uparrow$}} & \textit{\small{RO$\uparrow$}} & \textit{\small{VQ$\uparrow$}} & \textit{\small{\textbf{RS}$\uparrow$}} & \small{\textbf{Avg}} \\
\midrule

% ===================== Edit =====================
\multirow{7}{*}{\small{Edit}}
& \multirow{16}{*}{\textit{\small{w/o CoT}}}
& \small{FLUX.1-Kontext-dev~\cite{labs2025flux1kontextflowmatching}} & \textcolor{darkgreen}{\cmark} & -- & -- & -- & -- & -- &\small{\textbf{65.1}} & \small{13.1} & \small{\textbf{75.9}} & \small{1.6} & \small{\textbf{38.9}}\\
& & \small{FLUX.2-dev~\cite{flux-2-2025}} & \textcolor{darkgreen}{\cmark} & -- & -- & -- & -- & -- & \small{40.0} & \small{11.5} & \small{63.8} & \small{4.2} & \small{29.9} \\
& & \small{Qwen-Image-Edit-2509~\cite{wu2025qwenimagetechnicalreport}} & \textcolor{darkgreen}{\cmark} & -- & -- & -- & -- & -- & \small{52.4} & \small{\textbf{14.3}} & \small{60.7} & \small{1.4} & \small{32.2} \\
& & \small{Qwen-Image-Edit-2511~\cite{wu2025qwenimagetechnicalreport}} & \textcolor{darkgreen}{\cmark} & -- & -- & -- & -- & -- & \small{42.4} & \small{11.1} & \small{44.6} & \small{\textbf{4.9}} & \small{25.8} \\
& & \small{LongCat-Image-Edit~\cite{meituanlongcatteam2025longcatimagetechnicalreport}} & \textcolor{darkgreen}{\cmark} & -- & -- & -- & -- & -- & \small{53.6} & \small{11.4} & \small{60.5} & \small{3.3} & \small{32.2} \\
& & \small{Step1X-Edit~\cite{liu2025step1x-edit}} & \textcolor{darkgreen}{\cmark} & -- & -- & -- & -- & -- & \small{64.6} & \small{8.2} & \small{62.9} & \small{2.8} & \small{34.6} \\
& & \small{HiDream-E1.1~\cite{hidreami1technicalreport}} & \textcolor{darkgreen}{\cmark} & -- & -- & -- & -- & -- & \small{2.8} & \small{2.6} & \small{0.4} & \small{1.0} & \small{1.7} \\
& & \small{ICEdit~\cite{zhang2025icedit}} & \textcolor{darkgreen}{\cmark} & -- & -- & -- & -- & -- & \small{34.4} & \small{5.3} & \small{35.7} & \small{0.9} & \small{19.1} \\

% ===================== Unified (w/o CoT) =====================
\cmidrule(lr){1-1}\cmidrule(lr){3-14}
\multirow{13}{*}{\small{Unified}}
& & \small{Bagel~\cite{deng2025bagel}} & \textcolor{darkgreen}{\cmark} & -- & -- & -- & -- & -- & \small{35.5} & \small{8.7} & \small{46.0} & \small{2.2} & \small{23.1} \\
& & \small{OmniGen2~\cite{wu2025omnigen2explorationadvancedmultimodal}} & \textcolor{darkgreen}{\cmark} & -- & -- & -- & -- & -- & \small{27.7} & \small{6.3} & \small{54.6} & \small{0.8} & \small{22.4} \\
& & \small{UniWorld-V1~\cite{lin2025uniworldv1highresolutionsemanticencoders}} & \textcolor{darkgreen}{\cmark} & -- & -- & -- & -- & -- & \small{49.0} & \small{10.9} & \small{54.1} & \small{1.8} & \small{29.0} \\
& & \small{UniPic2-M-9B~\cite{wei2025skyworkunipic20building}} & \textcolor{darkgreen}{\cmark} & -- & -- & -- & -- & -- & \small{11.7} & \small{4.7} & \small{12.6} & \small{3.1} & \small{8.0} \\
& & \small{Ovis-U1-3B~\cite{wang2025ovisu1technicalreport}} & \textcolor{darkgreen}{\cmark} & -- & -- & -- & -- & -- & \small{42.0} & \small{7.0} & \small{46.9} & \small{1.2} & \small{24.3} \\
& & \small{DiMOO~\cite{xin2025dimoo}} & \textcolor{darkgreen}{\cmark} & -- & -- & -- & -- & -- & \textcolor{purple}{\small{\textbf{73.3}}} & \small{13.2} & \small{48.0} & \small{1.4} & \small{34.0} \\
& & \small{Seedream 4.0~\cite{seedream2025seedream40}} & \textcolor{red}{\xmark} & -- & -- & -- & -- & -- & \small{50.6} & \small{28.7} & \small{66.6} & \small{19.9} & \small{41.5} \\
& & \small{GPT-image-1~\cite{openai2024gpt-image-1}} & \textcolor{red}{\xmark} & -- & -- & -- & -- & -- & \small{57.5} & \small{29.3} & \small{87.6} & \small{13.4} & \small{46.9} \\
& & \small{Nano Banana~\cite{google2025nanobanana}} & \textcolor{red}{\xmark} & -- & -- & -- & -- & -- & \small{57.0} & \small{30.9} & \small{86.7} & \small{16.3} & \small{47.7} \\
& & \small{Nano Banana Pro~\cite{google2025nanobananapro}} & \textcolor{red}{\xmark} & -- & -- & -- & -- & -- & \small{70.2} & \textcolor{purple}{\small{\textbf{62.0}}} & \textcolor{purple}{\small{\textbf{95.1}}} & \textcolor{purple}{\small{\textbf{46.4}}} & \textcolor{purple}{\small{\textbf{68.4}}} \\

% ===================== Unified (w/ CoT) =====================
\cmidrule(lr){2-14}
& \multirow{6}{*}{\textit{\small{w/ CoT}}}
& \small{Bagel-Think~\cite{deng2025bagel}} & \textcolor{darkgreen}{\cmark} & \small{15.2} & \small{4.5} & \small{36.5} & \small{2.2} & \small{14.6} & \small{8.2} & \small{6.1} & \small{21.3} & \small{2.3} & \small{9.5} \\
& & \small{Zebra-CoT~\cite{li2025zebracot}} & \textcolor{darkgreen}{\cmark} & \small{49.9} & \small{8.9} & \small{57.6} & \small{2.7} & \small{29.8} & \small{42.1} & \small{13.3} & \small{35.4} & \small{1.6} & \small{23.1} \\
& & \small{Uni-CoT~\cite{qin2025unicot}} & \textcolor{darkgreen}{\cmark} & \small{34.2} & \small{10.1} & \small{47.6} & \small{5.3} & \small{24.3} & \small{26.3} & \small{9.3} & \small{33.3} & \small{3.1} & \small{18.0} \\
& & \small{GPT-image-1\textsuperscript{\dag}~\cite{openai2024gpt-image-1}} & \textcolor{red}{\xmark} & \small{67.4} & \small{35.8} & \small{80.7} & \small{32.9} & \small{54.2} & \small{27.8} & \small{25.7} & \small{52.0} & \small{19.7} & \small{31.3} \\
& & \small{Nano Banana\textsuperscript{\dag}~\cite{google2025nanobanana}} & \textcolor{red}{\xmark} & \small{82.2} & \small{40.7} & \textcolor{purple}{\small{\textbf{92.6}}} & \small{37.7} & \small{63.3} & \small{63.8} & \small{36.8} & \small{79.1} & \small{22.3} & \small{50.5} \\
& & \small{Nano Banana Pro\textsuperscript{\dag}~\cite{google2025nanobananapro}} & \textcolor{red}{\xmark} & \textcolor{purple}{\small{\textbf{86.0}}} & \textcolor{purple}{\small{\textbf{58.6}}} & \small{90.9} & \textcolor{purple}{\small{\textbf{52.0}}} & \textcolor{purple}{\small{\textbf{72.0}}} & \small{\textbf{66.3}} & \small{\textbf{54.5}} & \small{\textbf{83.9}} & \small{\textbf{40.2}} & \small{\textbf{61.2}} \\

% ===================== Video =====================
\midrule
\multirow{5}{*}{\small{Video Gen}}
& \multirow{5}{*}{\textit{\small{Video}}}
& \small{Wan 2.2~\cite{wan2025}} & \textcolor{darkgreen}{\cmark} & \small{61.5} & \small{11.9} & \small{67.0} & \small{7.4} & \small{37.0} & \small{31.2} & \small{6.4} & \small{36.7} & \small{1.1} & \small{18.9} \\
& & \small{Kling 1.6~\cite{kuaishou2024kling}} & \textcolor{red}{\xmark} & \small{\textbf{73.9}} & \small{12.4} & \small{77.0} & \small{9.6} & \small{43.2} & \small{\textbf{59.5}} & \small{6.3} & \small{\textbf{52.5}} & \small{1.6} & \small{\textbf{30.0}} \\
& & \small{Seedance 1.0 Pro~\cite{bytedance2025seedance}} & \textcolor{red}{\xmark} & \small{39.5} & \small{12.1} & \small{63.5} & \small{9.2} & \small{31.1} & \small{48.0} & \small{14.0} & \small{51.5} & \small{4.8} & \small{29.6} \\
& & \small{Veo 3~\cite{googledeepmind2024veo3}} & \textcolor{red}{\xmark} & \small{69.6} & \small{36.2} & \small{85.3} & \small{31.9} & \small{55.8} & \small{33.6} & \small{15.0} & \small{25.0} & \small{8.4} & \small{20.5} \\
& & \small{Sora 2 Pro~\cite{openai2025sora2}} & \textcolor{red}{\xmark} & \small{70.5} & \small{\textbf{38.8}} & \small{\textbf{85.5}} & \small{\textbf{34.8}} & \small{\textbf{57.4}} & \small{24.5} & \small{\textbf{16.6}} & \small{20.0} & \small{\textbf{10.1}} & \small{17.8} \\

\bottomrule
\end{tabular}
}
% \vspace{-0.2in}
\end{table*}

%>>>>>>>>>>>>>>>>>>>>>>>>>Main Result>>>>>>>>>>>>>>>>>>>>>>>>

\textbf{Experimental Setup.} We constructed a random ``tiny split" from our benchmark generation results, comprising 1,080 final result outputs (yielding 4,272 metric evaluation instances) and 540 process sequences (yielding 1,064 metric evaluation instances). Three human experts independently scored these instances using the same input information and templates as the VLM. The average of the three human scores serves as the ``Gold Standard." Simultaneously, Gemini-2.5-Pro evaluated the same set over three independent runs under two settings: \textit{with} Ground Truth (GT) references and \textit{without} them.

\textbf{Metrics.} We assess reliability via three dimensions: (1) MAE (Mean Absolute Error), measuring the distributional distance between the VLM's average score and the human average; (2) Accuracy, measuring categorical agreement. For the continuous Process Metrics (0--100), scores were discretized into three intervals—Bad [0, 33], Moderate [34, 67], and Good [68, 100]—to calculate alignment; (3) Variance, quantifying the stability and consistency of the evaluator across three runs.

\textbf{Analysis.} As presented in Table~\ref{tab:evaluation-results}, the results support three key conclusions:
\begin{itemize}[leftmargin=*, noitemsep, topsep=2pt]
    \item \textbf{High Human Alignment:} When provided with GT references, Gemini-2.5-Pro achieves high accuracy (73.3\% for Process, 78.6\% for Result) and low MAE, demonstrating that the VLM's judgment distribution closely approximates that of human experts.
    \item \textbf{Criticality of Ground Truth:} There is a significant performance gap between the \textit{w/} and \textit{w/o} GT settings. The inclusion of GT references substantially reduces MAE and Variance, confirming that providing golden references is essential for stabilizing VLM judgments.
    \item \textbf{Stability:} The variance of the VLM is comparable to, and in some cases lower than, the inter-annotator variance of human experts. This indicates that our automated pipeline offers a consistency level competitive with human consensus, making it a reliable proxy for large-scale evaluation.
\end{itemize}

% \begin{figure*}[t]
%     \centering
%     \includegraphics[width=\textwidth]{figures/examples.pdf}
%     \caption{\textbf{Overview of the ViGoR-Bench task suite.} We present representative demo cases and their corresponding editing instructions across 20 distinct sub-tasks. These tasks are hierarchically organized into three primary reasoning domains: Physical Reasoning, Knowledge Reasoning, and Symbolic Reasoning.}
%     \label{fig:demo}
% \end{figure*}

% \subsection{Implementation Details}
% \label{sec:implementation_details}

\section{Experiment}

%>>>>>>>>>>>>>>>>>>>>>>>>Qualitative exp>>>>>>>>>>>>>>>>>>>>>>>>
\begin{figure*}[t!]
    \centering
    \includegraphics[width=1.0\textwidth]{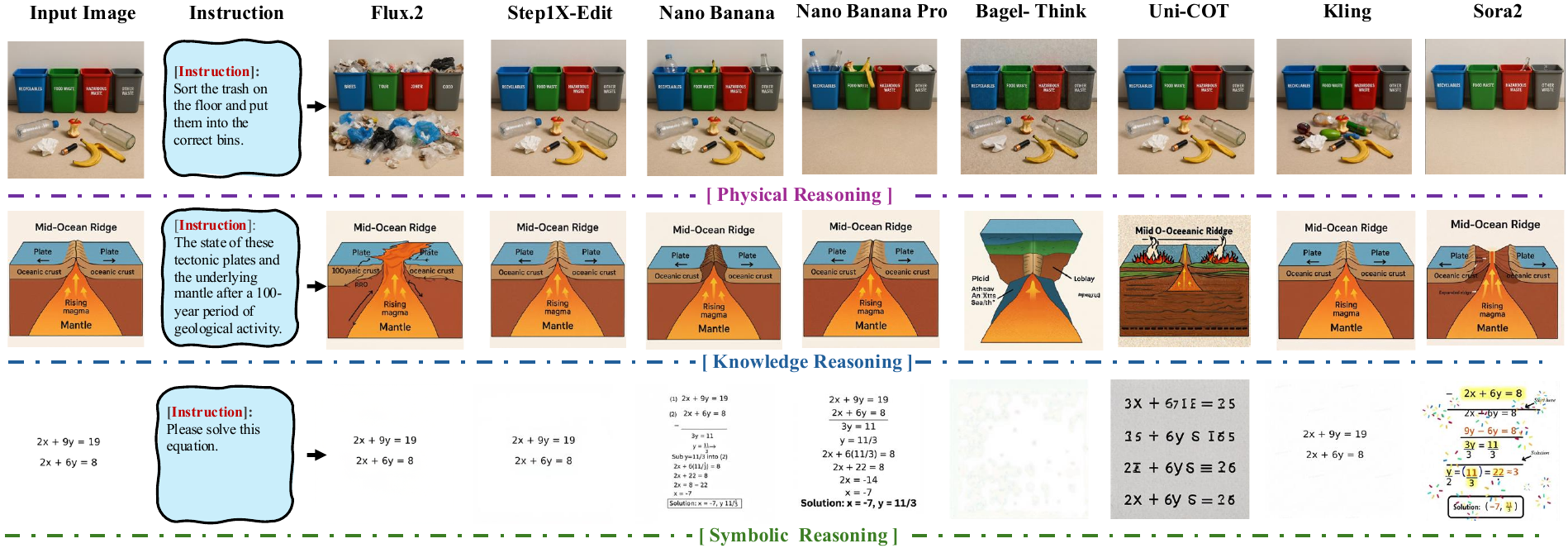}
    \caption{\textbf{Qualitative comparison of leading models.} We present case studies across three representative reasoning domains.}
    \label{fig:Qualitative}
    \vspace{-0.3in}
\end{figure*}
%>>>>>>>>>>>>>>>>>>>>>>>>Qualitative exp>>>>>>>>>>>>>>>>>>>>>>>>

\textbf{Implementation Details.} We conduct a comprehensive evaluation of leading open-source and proprietary models, categorized into four distinct groups: \textbf{Image Editing Models}, \textbf{Unified Models without Chain-of-Thought}, \textbf{Unified Models with CoT}, and \textbf{Video Generation Models}. To strictly assess generalization in a zero-shot setting, we utilize official checkpoints and standard APIs, adhering to the default inference parameters recommended by their respective official implementations to ensure a fair comparison. For clarity and consistency across diverse metrics, all reported scores are normalized to a 100-point scale.

\subsection{Main Results}
\label{sec:main_results}

Table~\ref{tab:main_results} presents the quantitative performance of leading models across our proposed metrics. Complementing this, Figure~\ref{fig:Qualitative} provides a qualitative comparison of representative cases. Our analysis yields three critical observations regarding the current state of visual reasoning.

\textbf{Proprietary models maintain a significant performance lead over their open-source counterparts.}
Proprietary unified models continue to maintain a substantial lead over open-source counterparts. Nano Banana Pro consistently secures the top performance across most metrics. This quantitative gap is visually corroborated in Figure~\ref{fig:Qualitative}. In complex domains such as \textit{Physical Reasoning} and \textit{Symbolic Reasoning}, only top-tier models like Nano Banana Pro and Sora 2 Pro demonstrate the capacity to generate accurate results. In contrast, other models, such as Flux.2~\cite{flux-2-2025} and Bagel-Think~\cite{deng2025bagel}, frequently exhibit hallucinations or fail to adhere to the specified constraints, highlighting the persistent challenge of instruction following in complex visual reasoning scenarios.

\textbf{Explicit CoT prompting enhances the interpretability of the generation process but it does not guarantee an improvement in final accuracy.}
We investigate the impact of explicit reasoning steps on visual generation. Notably, models marked with \textsuperscript{\dag} (e.g., GPT-image-1\textsuperscript{\dag}~\cite{openai2024gpt-image-1}, Nano Banana Pro\textsuperscript{\dag}~\cite{google2025nanobananapro}) lack native interleaved generation capabilities; thus, we employed external planners (GPT-5~\cite{singh2025openaigpt5card}/Gemini-2.5-Pro~\cite{gemini25pushingfrontier}) to decompose tasks into sequential generation steps.
While CoT significantly enhances the interpretability of the intermediate process and ensures logical chain completeness, it does not strictly guarantee superior final outcomes. Task decomposition aids in clarifying the reasoning trajectory; however, it does not necessarily compensate for the base model's execution limitations—effectively, a model may ``think" correctly but fail to ``draw" accurately. Therefore, the introduction of CoT does not automatically translate to improved Result Metrics, which demand high-fidelity visual grounding. Moreover, the elongation of the inference chain introduces the risk of \textit{error accumulation}, where minor execution deviations in early steps cascade into compounded failures in the final output.

\textbf{Video generation models often exhibit an ``Illusion of Reasoning'', where apparent logical consistency does not hold up to rigorous evaluation.} 
As shown in Table~\ref{tab:main_results}, video generation models (e.g., Kling 1.6~\cite{kuaishou2024kling}, Sora 2 Pro~\cite{openai2025sora2}) demonstrate exceptional temporal consistency, achieving Process Visual Quality (VQ) scores (e.g., 77.0\% and 85.5\%) that are comparable to, or even exceed, those of top-tier \textit{Unified w/ CoT} models. However, a stark contrast exists in their logical efficacy: their Result Reasoning Success (RS) scores remain disproportionately low (e.g., 1.6\% for Kling 1.6). This discrepancy suggests that current video models excel at simulating fluid motion and maintaining visual coherence but struggle to internalize the underlying logical constraints required for rigorous reasoning tasks.

\subsection{Analysis}
\label{sec:analysis}

%Beyond aggregate rankings, we conduct a fine-grained diagnosis of model capabilities.

% \textbf{Capability Profiling.} 
% Figure~\ref{fig:radar_puzzle} illustrates the performance variance across symbolic sub-tasks. Leading models demonstrate high proficiency in \textit{Algebraic Calculation} but exhibit significant performance deficits in \textit{Jigsaw Puzzle} and \textit{Function Plotting}, highlighting clear areas for improvement. Comprehensive visualizations for the Physical and Knowledge Reasoning domains are provided in the Appendix.

\begin{figure}[t]
    \centering
    \includegraphics[width=1.0\columnwidth]{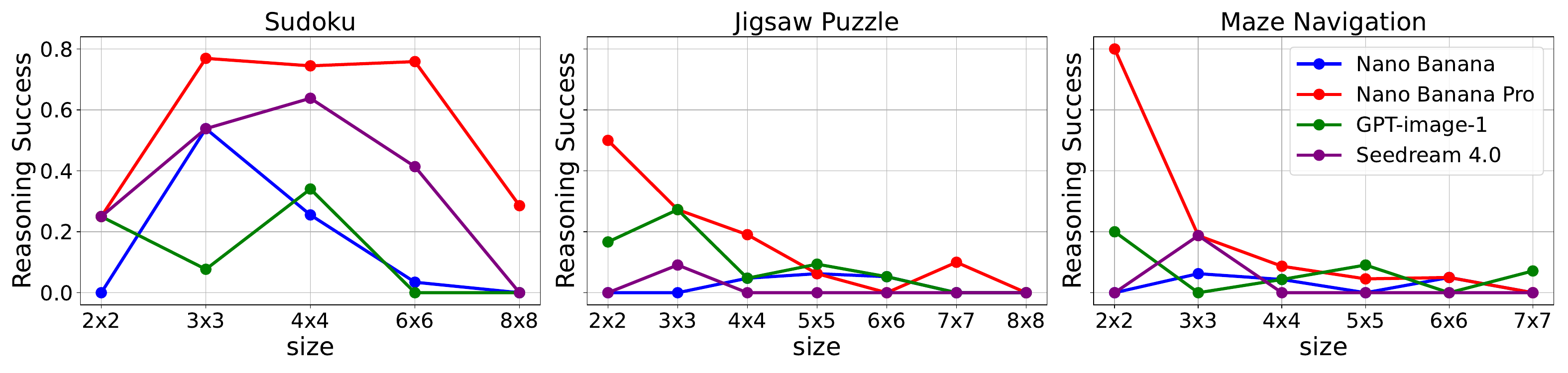}
    \caption{\textbf{Impact of problem complexity on Reasoning Success.} We report the performance of evaluated models on Sudoku, Jigsaw Puzzle, and Maze Navigation tasks across varying grid dimensions.}

    \label{fig:line_chart}
    \vspace{-0.4in}
\end{figure}

\textbf{Impact of Problem Complexity.} 
Figure~\ref{fig:line_chart} investigates whether model performance mimics human-like degradation as problem dimensionality increases. For \textit{Maze Navigation} and \textit{Jigsaw Puzzle}, we observe a sharp, monotonic decline in Reasoning Success as the grid size expands. However, \textit{Sudoku} presents an intriguing inverted-U pattern: performance peaks at intermediate dimensions but drops at both extremes. We hypothesize this stems from training data distribution biases, where standard grid sizes are over-represented compared to non-standard variants.

\subsection{Eliciting Reasoning via Post-training}
\label{sec:post_training}

Finally, to demonstrate the practical utility of ViGoR-Bench in guiding model improvement of reasoning, we investigate whether our benchmark data and metrics can serve as effective signals for training.

\textbf{Setup.} We constructed three distinct training sets, each containing 10k synthetic \textit{Maze Navigation} samples with grid dimensions of $4\times4$, $6\times6$, and $8\times8$, respectively. Using Qwen-Image-Edit (versions 2509 and 2511) as base models, we applied Supervised Fine-Tuning (SFT) followed by Reinforcement Learning (RL) using the GRPO~\cite{liu2025flow} algorithm. Crucially, all fine-tuned models were evaluated on maze navigation sub-domain in our benchmark, which spans grid dimensions from $2\times2$ to $7\times7$.

\textbf{Surpassing SOTA Proprietary Models.} 
As detailed in Table~\ref{tab:post-train}, post-training successfully elicits reasoning capabilities. The performance gains are substantial: specifically, the Qwen-Image-Edit-2511-RL model trained on $8\times8$ data achieves a remarkable \textbf{Reasoning Success (RS) of 97.0\%} and an \textbf{Average Score of 99.0}. This result not only represents a quantum leap over its base model but also convincingly outperforms the state-of-the-art proprietary model.

\textbf{Training on more challenging data enhances a model's generalization performance on simpler visual reasoning tasks.} 
A comparison across training splits reveals a compelling insight into generalization. While models trained on in-domain distributions ($4\times4$ and $6\times6$) show solid improvements, the model trained on the strictly Out-Of-Distribution (OOD) and higher-complexity $8\times8$ data yields the best overall performance. Despite the $8\times8$ grid being more difficult than any case in the test set, learning from these ``harder" examples fosters robust logic that transfers effectively to easier tasks. This suggests that training on high-complexity data forces the model to learn the underlying reasoning rules rather than merely overfitting to surface patterns.

\begin{figure}[h] % 建议用 [t] 让图片置顶，[h] 在双栏排版中有时会乱跑
    \centering
    \begin{subfigure}[b]{0.45\linewidth}
        \centering
        \includegraphics[width=\linewidth]{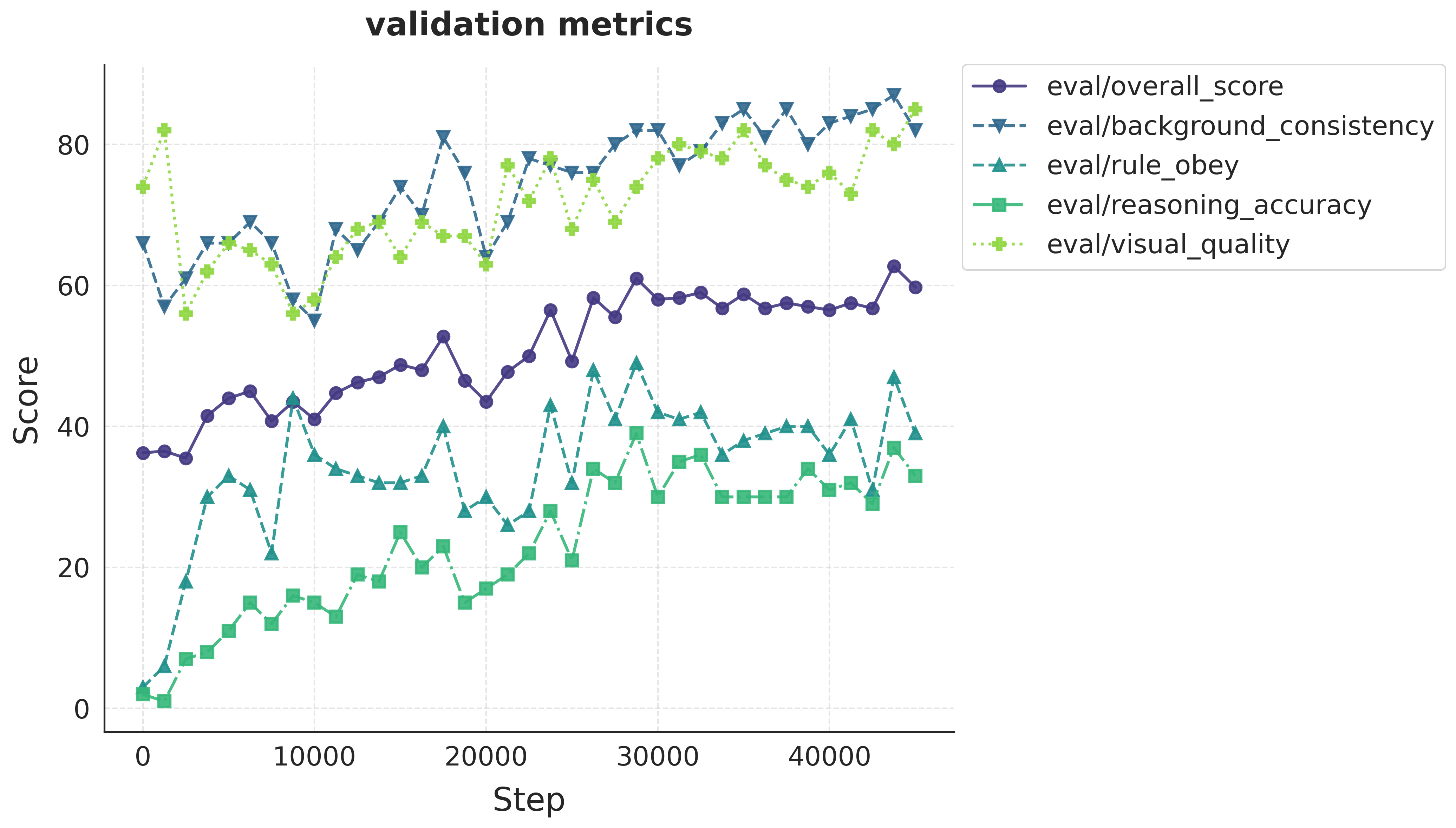}
        \caption{SFT}
        \label{fig:img2}
    \end{subfigure}
    \hfill
    \begin{subfigure}[b]{0.45\linewidth}
        \centering
        \includegraphics[width=\linewidth]{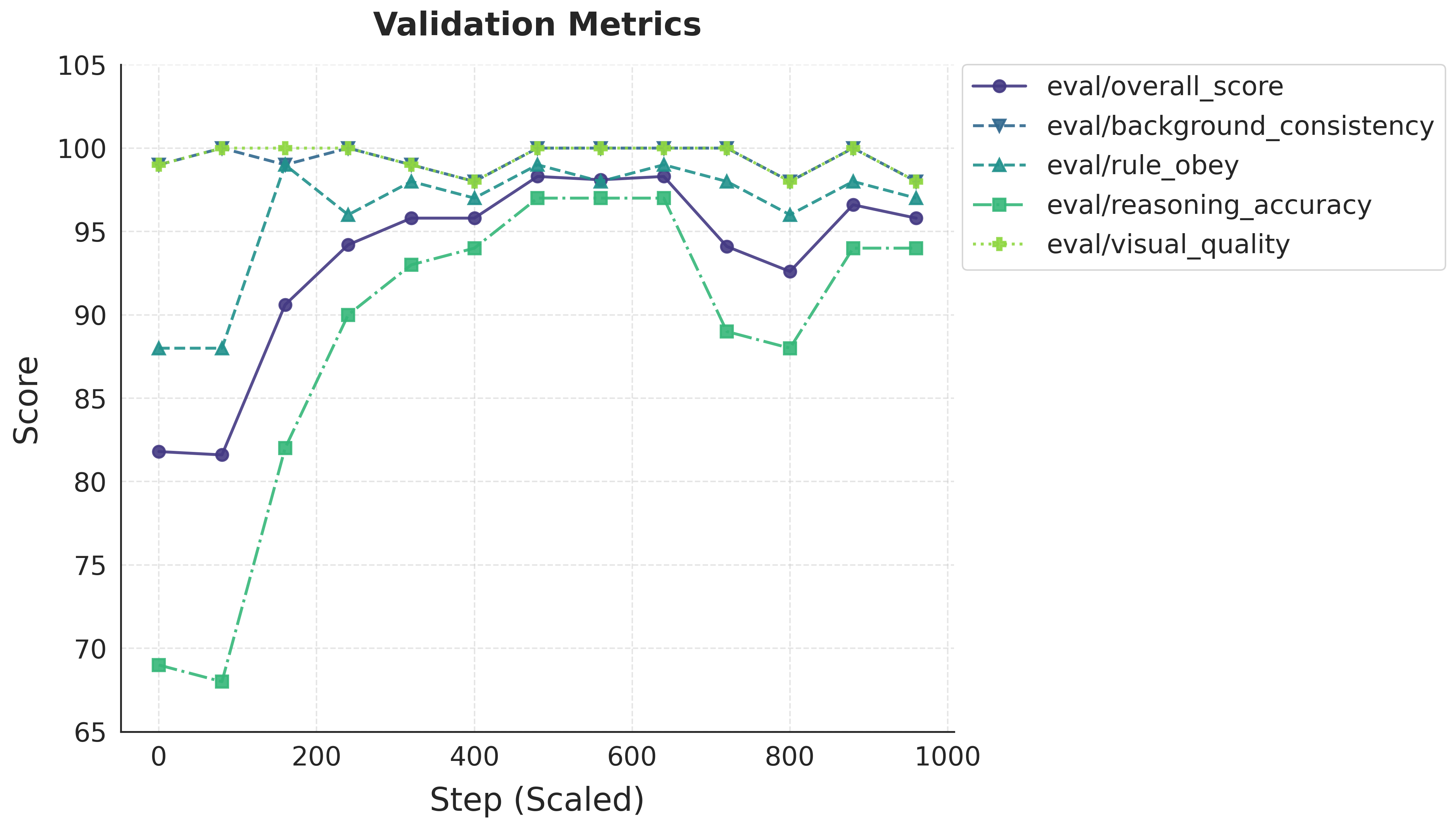}
        \caption{RL}
        \label{fig:img4}
    \end{subfigure}
    
    \caption{Overview of results of SFT and RL}
    \label{fig:four_grid}
    \vspace{-0.15in}
\end{figure}

\textbf{Reward-driven RL demonstrates superior potential in advancing visual reasoning capabilities where SFT exhibits saturation.}
As illustrated in the Figure~\ref{fig:four_grid}, the performance of the Supervised Fine-Tuning (SFT) model, as measured by validation metrics, reaches a plateau during the course of training. However, by leveraging the SFT model as a foundation, an additional phase of Reinforcement Learning (RL) training successfully elevates the model's performance to a new level.

\begin{table}[t]
\centering
\caption{\textbf{Comparison of SFT and RL fine-tuning stages.} Models are fine-tuned on constructed maze datasets with grid dimensions of $4\times4$, $6\times6$, and $8\times8$. Evaluation is conducted on our benchmark test set, which covers maze grid dimensions ranging from $2\times2$ to $7\times7$.}
\label{tab:post-train}
\tablestyle{5.0pt}{1.0}
\resizebox{\columnwidth}{!}{%
    \begin{tabular}{l|c|ccccc}
    \toprule
    % 【修正】原代码中的 \samll 修正为 \small
    Model & Maze Grid & \textit{\small{BC~$\uparrow$}} & \textit{\small{RO~$\uparrow$}} & \textit{\small{VQ~$\uparrow$}} & \textit{\small{\textbf{RS}~$\uparrow$}} & \textbf{Avg} \\
    \midrule
    
    % --- 第一大栏：Baselines ---
    \cellcolor{gray!10}\small{Qwen-Image-Edit-2509} & \multirow{5}{*}{\small{N/A}} & \cellcolor{gray!10}\small{0.0} & \cellcolor{gray!10}\small{0.0} & \cellcolor{gray!10}\small{32.0} & \cellcolor{gray!10}\small{0.0} & \cellcolor{gray!10}\small{8.0} \\
    \cellcolor{gray!10}\small{Qwen-Image-Edit-2511} & & \cellcolor{gray!10}\small{66.0} & \cellcolor{gray!10}\small{3.0} & \cellcolor{gray!10}\small{74.0} & \cellcolor{gray!10}\small{2.0} & \cellcolor{gray!10}\small{36.3} \\
    \small{Nano Banana} & & \small{38.0} & \small{37.0} & \small{96.0} & \small{3.0} & \small{43.5} \\
    \small{GPT-Image-1} & & \small{96.0} & \small{23.0} & \small{98.0} & \small{5.0} & \small{55.5} \\
    \small{Nano Banana Pro} & & \small{93.0} & \small{50.0} & \small{98.0} & \small{11.0} & \small{63.0} \\
    \midrule
    
    % --- 第二大栏：4x4 数据 ---
    \small{Qwen-Image-Edit-2509-SFT} & \multirow{4}{*}{\small{$4\times4$}} & \small{65.0} & \small{65.0} & \small{68.0} & \small{27.0} & \small{56.3} \\
    % 2509-RL 用淡蓝色 (azure)
    \cellcolor{azure}\small{Qwen-Image-Edit-2509-RL} & & \cellcolor{azure}\textbf{\small{100.0}} & \cellcolor{azure}\small{69.0} & \cellcolor{azure}\small{96.0} & \cellcolor{azure}\small{60.0} & \cellcolor{azure}\small{81.3} \\
    \small{Qwen-Image-Edit-2511-SFT} & & \small{82.0} & \small{11.0} & \small{65.0} & \small{11.0} & \small{42.3} \\
    % 2511-RL 用淡紫色 (lavender)
    \cellcolor{lavender}\small{Qwen-Image-Edit-2511-RL} & & \cellcolor{lavender}\textbf{\small{100.0}} & \cellcolor{lavender}\small{67.0} & \cellcolor{lavender}\small{97.0} & \cellcolor{lavender}\small{59.0} & \cellcolor{lavender}\small{80.8} \\
    \midrule
    
    % --- 第三大栏：6x6 数据 ---
    \small{Qwen-Image-Edit-2509-SFT} & \multirow{4}{*}{\small{$6\times6$}} & \small{62.0} & \small{87.0} & \small{72.0} & \small{42.0} & \small{65.8} \\
    % 2509-RL 用淡蓝色
    \cellcolor{azure}\small{Qwen-Image-Edit-2509-RL} & & \cellcolor{azure}\textbf{\small{100.0}} & \cellcolor{azure}\small{82.0} & \cellcolor{azure}\small{99.0} & \cellcolor{azure}\small{80.0} & \cellcolor{azure}\small{90.3} \\
    \small{Qwen-Image-Edit-2511-SFT} & & \small{63.0} & \small{52.0} & \small{46.0} & \small{43.0} & \small{51.0} \\
    % 2511-RL 用淡紫色
    \cellcolor{lavender}\small{Qwen-Image-Edit-2511-RL} & & \cellcolor{lavender}\textbf{\small{100.0}} & \cellcolor{lavender}\small{94.0} & \cellcolor{lavender}\small{97.0} & \cellcolor{lavender}\small{81.0} & \cellcolor{lavender}\small{93.0} \\
    \midrule
    
    % --- 第四大栏：8x8 数据 ---
    \small{Qwen-Image-Edit-2509-SFT} & \multirow{4}{*}{\small{$8\times8$}} & \small{57.0} & \small{86.0} & \small{74.0} & \small{39.0} & \small{64.0} \\
    % 2509-RL 用淡蓝色
    \cellcolor{azure}\small{Qwen-Image-Edit-2509-RL} & & \cellcolor{azure}\textbf{\small{100.0}} & \cellcolor{azure}\small{84.0} & \cellcolor{azure}\small{98.0} & \cellcolor{azure}\small{72.0} & \cellcolor{azure}\small{88.5} \\
    \small{Qwen-Image-Edit-2511-SFT} & & \small{82.0} & \small{49.0} & \small{74.0} & \small{39.0} & \small{61.0} \\
    % 2511-RL 用淡紫色
    \cellcolor{lavender}\small{Qwen-Image-Edit-2511-RL} & & \cellcolor{lavender}\textbf{\small{100.0}} & \cellcolor{lavender}\textbf{\small{99.0}} & \cellcolor{lavender}\textbf{\small{100.0}} & \cellcolor{lavender}\textbf{\small{97.0}} & \cellcolor{lavender}\textbf{\small{99.0}} \\
    
    \bottomrule
    \end{tabular}%
}
\vspace{-0.15in}
\end{table}
%>>>>>>>>>>>>>>>>>>>>>>>>Post-train exp>>>>>>>>>>>>>>>>>>>>>>>>
\section{Conclusion}
\label{sec:conclusion}

In this paper, we introduced \textbf{ViGoR-Bench}, a comprehensive benchmark coupled with a rigorous dual-track evaluation protocol designed to assess diverse visual generation models. We empirically validated the reliability of our automated evaluation pipeline against human experts. Through extensive experiments, we quantified the current limitations of state-of-the-art models, particularly their performance degradation on high-complexity puzzle tasks.  

%Furthermore, our post-training experiments provide a critical insight: visual reasoning capabilities can be effectively elicited via supervised fine-tuning and reinforcement learning. Notably, models trained on high-difficulty data demonstrated strong generalization to easier, in-domain tasks, surpassing proprietary baselines. We hope this work serves as a foundational step for future research, particularly in developing scaling laws for visual reasoning and exploring unified training paradigms that intrinsically integrate logical planning with visual synthesis.

\clearpage
\section*{Impact Statement}
The goal of ViGoR-Bench is to catalyze progress in generative AI by shifting the focus from mere visual fidelity to genuine reasoning capabilities. By enabling researchers to identify and address logical deficits in current models, our work can accelerate the development of more reliable and intelligent AI systems. A direct positive impact is the potential for safer and more dependable AI. Models that better understand physical laws and causal reasoning are less likely to generate nonsensical or harmful content, making them more suitable for critical applications in fields like education, scientific simulation, and engineering design. our benchmark ultimately contributes to a more responsible and transparent AI ecosystem. We encourage the community to use ViGoR-Bench not only for model development but also for research into detecting and mitigating potential risks.

\bibliography{references}
\bibliographystyle{icml2026}

\clearpage
% \section{Appendix}
% \subsection{Dataset Statistics and Qualitative Examples}

% \newpage
\appendix
\onecolumn
\section{Dataset Statistics and Qualitative Examples}

% \begin{figure*}[h]
%     \centering
%     \includegraphics[width=\textwidth]{figures/statistic.pdf}
%     \caption{Statistic of ViGoR-Bench.}
%     \label{fig:statistic}
% \end{figure*}

\begin{figure*}[t]
    \centering
    \includegraphics[width=\textwidth]{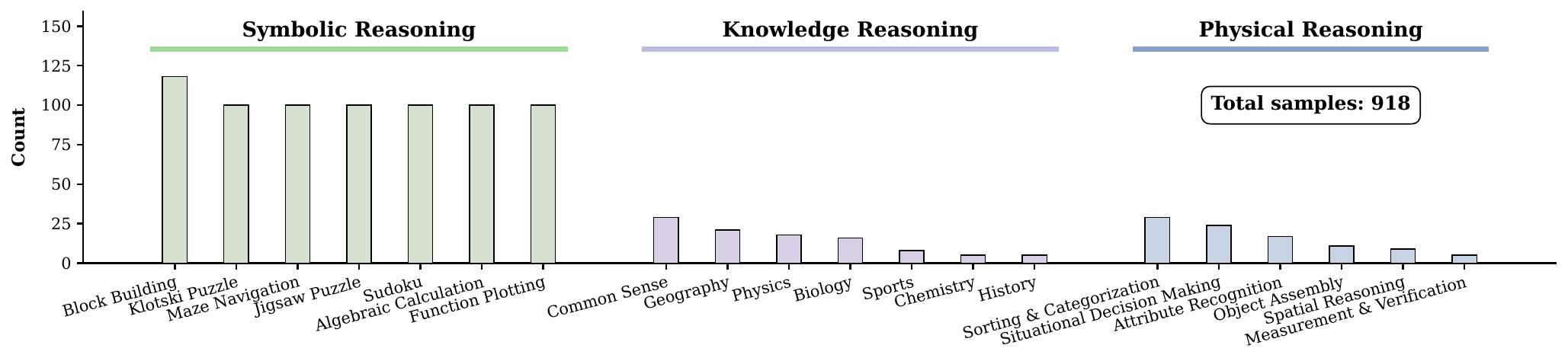}
    \caption{Statistic of ViGoR-Bench.}
    \label{fig:statistic}
\end{figure*}

Figure~\ref{fig:statistic} presents the statistical breakdown of ViGoR-Bench.
The benchmark consists of 918 samples spanning three major reasoning categories:
\emph{Symbolic Reasoning}, \emph{Knowledge Reasoning}, and \emph{Physical Reasoning}.

\begin{figure*}[t]
    \centering
    \includegraphics[width=\textwidth]{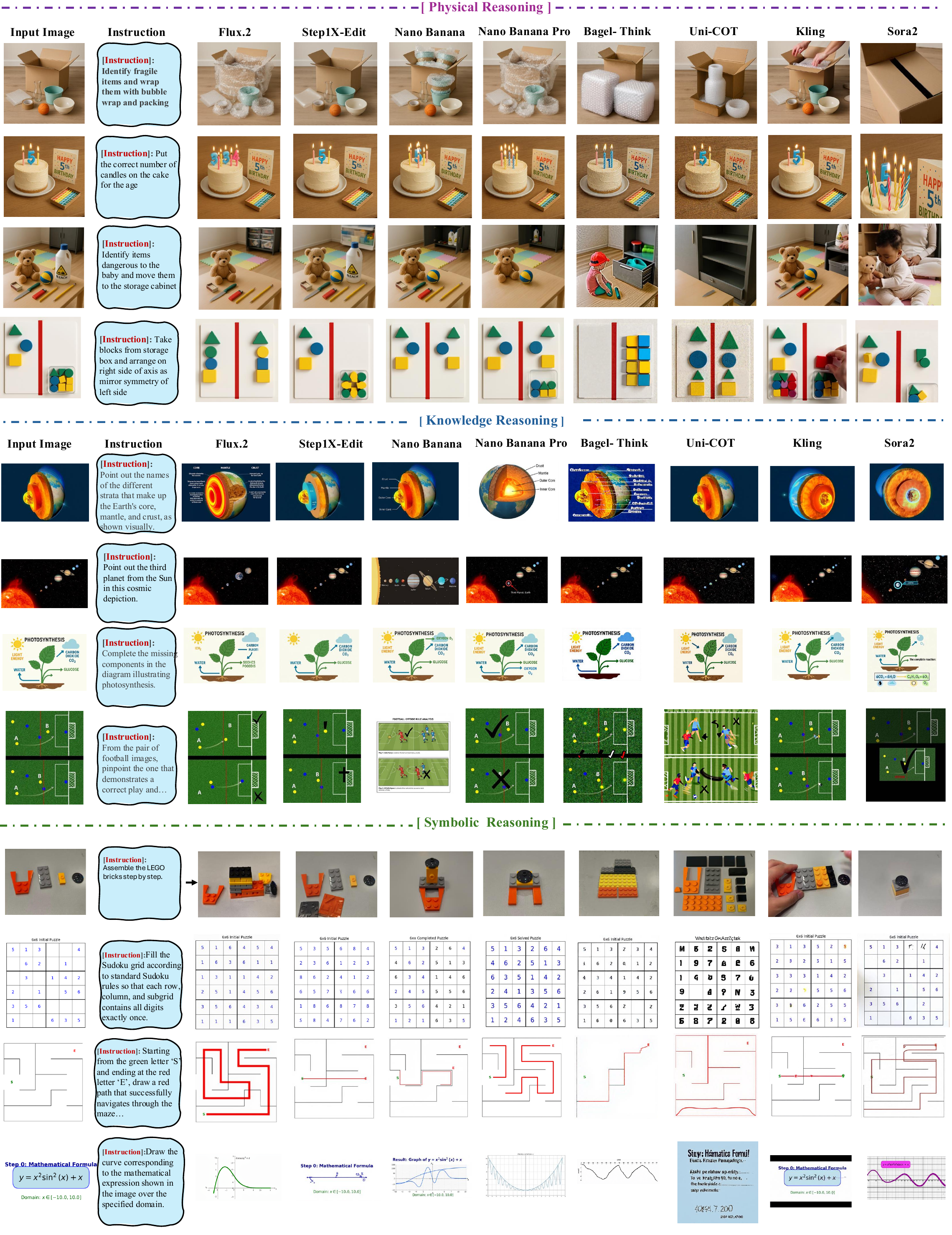}
    \caption{Samples of ViGoR-Bench.}
    \label{fig:appendix_cases}
\end{figure*}

Figure~\ref{fig:appendix_cases} illustrates representative samples from ViGoR-Bench.
Each example pairs an instruction with a sequence of intermediate visual states, highlighting the emphasis on process-aware reasoning rather than final outcomes alone.
The samples demonstrate diverse reasoning patterns, including symbolic constraint satisfaction (e.g., Sudoku completion), path planning and navigation, and mathematical function construction, showcasing the benchmark’s ability to evaluate multi-step visual reasoning processes.

\begin{figure*}[h]
    \centering
    \includegraphics[width=\textwidth]{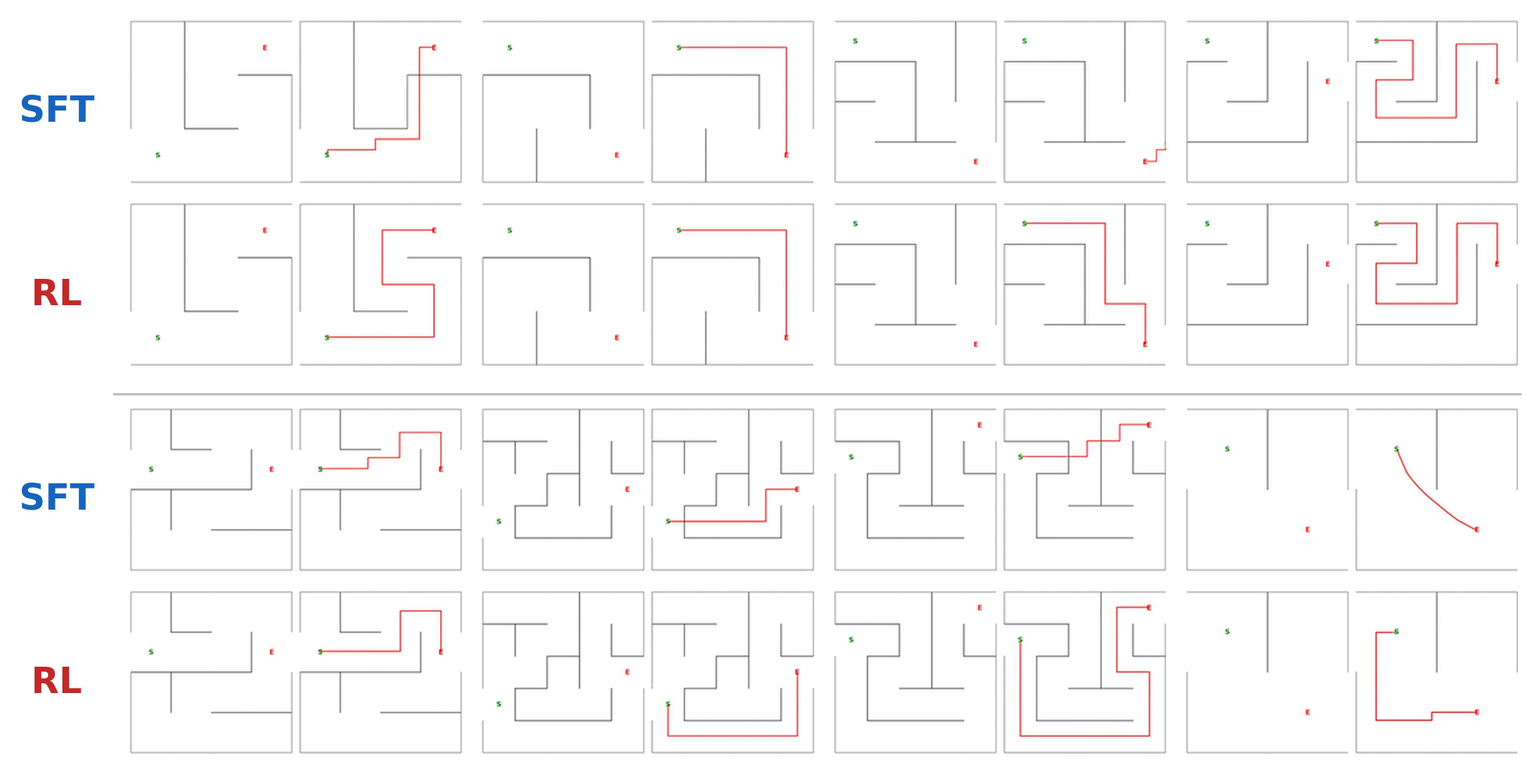}
    \caption{Comparison of qualitative results for Qwen-Image-Edit-2511 on ViGoR-Bench after SFT and RL.}
    \label{fig:sft_rl}
\end{figure*}

Figure~\ref{fig:sft_rl} presents a qualitative comparison of the Qwen-Image-Edit-2511~\cite{wu2025qwenimagetechnicalreport} model's performance on ViGoR-Bench following SFT and RL. The visual trajectories demonstrate that compared to relying solely on SFT, the integration of RL enhances the model's capability to solve complex problems, particularly in high-dimensional mazes. Specifically, the post-RL model exhibits a higher probability of successful reasoning, with marked improvements across three key dimensions: background consistency, rule obey (e.g., strictly avoiding wall collisions), and reasoning success.

\section{Capability Profiling.}

\begin{figure*}[t]
    \centering
    \includegraphics[width=\textwidth]{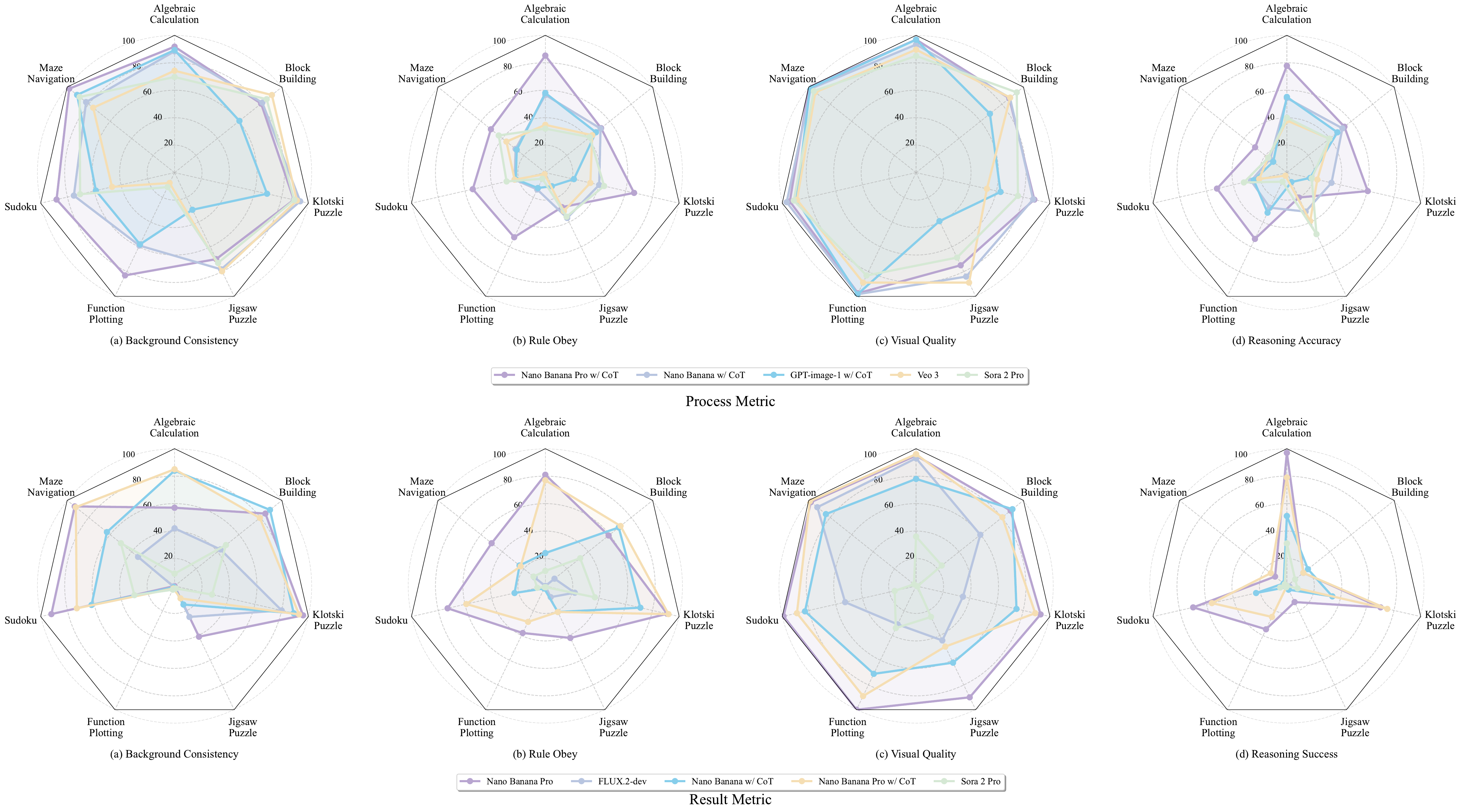}
    \caption{\textbf{Performance profiling on symbolic reasoning tasks.} Comparison of five models across seven sub-tasks. The top and bottom rows report performance on four Process Metrics and four Result Metrics.}
    \label{fig:radar_puzzle}
    \vspace{-0.25in}
\end{figure*}

Figure~\ref{fig:radar_puzzle} illustrates the performance variance across symbolic reasoning sub-tasks.
Leading models demonstrate consistently strong performance in \emph{Algebraic Calculation} and \emph{Block Building}, achieving high scores in both process-level and result-level metrics.
In contrast, substantially lower performance is observed for combinatorial and structural tasks such as \emph{Jigsaw Puzzle}, \emph{Function Plotting}, and \emph{Maze Navigation}, highlighting clear limitations in handling multi-step symbolic manipulation and spatially structured reasoning.

Across process metrics, models generally maintain high \emph{Background Consistency} and \emph{Visual Quality}, whereas pronounced drops are evident in \emph{Rule Obey}, particularly for tasks requiring strict symbolic constraints.
Result-level evaluations further amplify this gap, with \emph{Reasoning Accuracy} and \emph{Reasoning Success} exhibiting significant degradation on puzzle-oriented tasks, indicating that visually plausible intermediate states do not reliably translate into correct symbolic reasoning outcomes.

\begin{figure*}[h!]
    \centering
    \includegraphics[width=\textwidth]{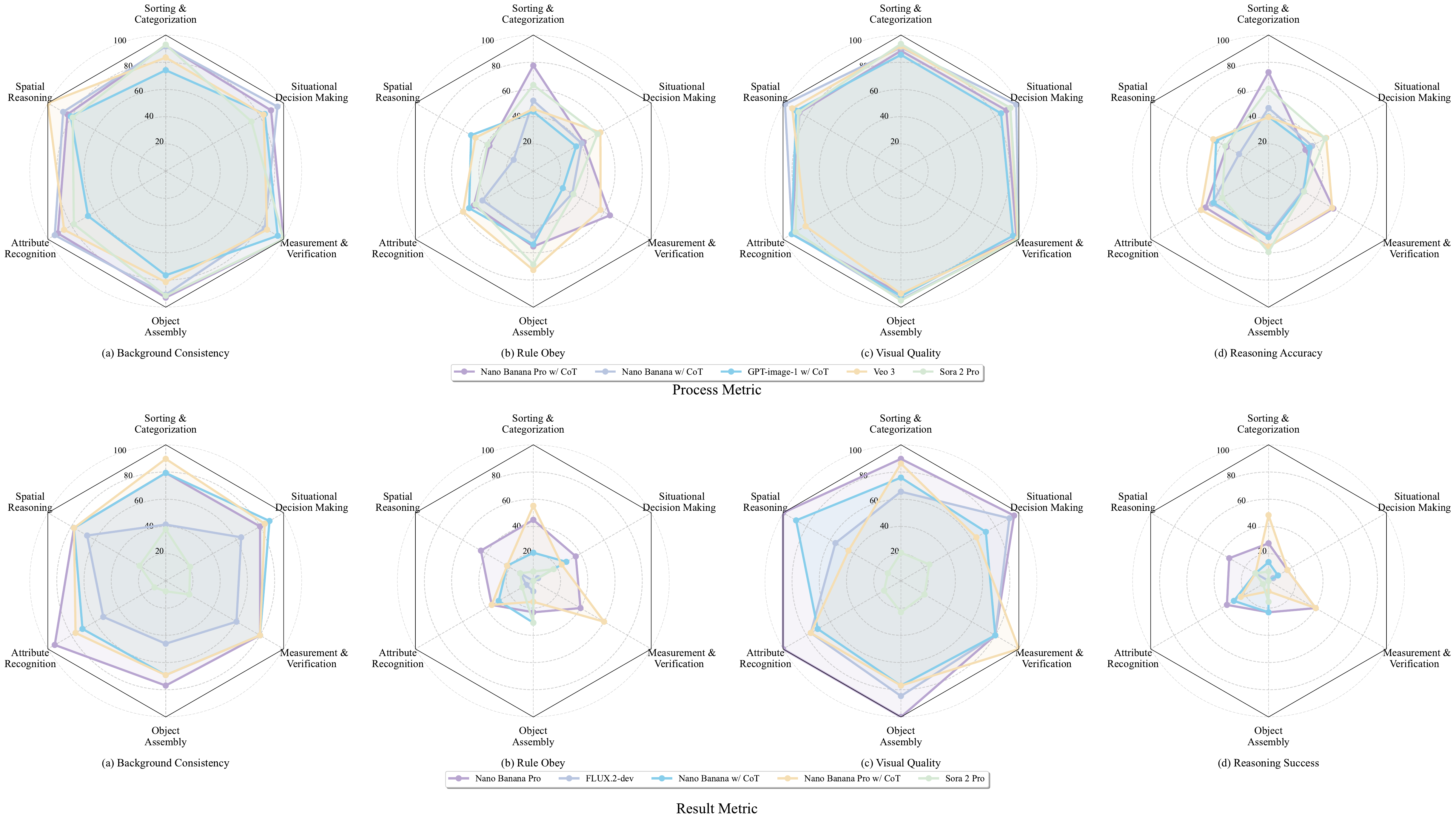}
    \caption{\textbf{Performance profiling on physical reasoning tasks.} Comparison of five models across seven sub-tasks. The top and bottom rows report performance on four Process Metrics and four Result Metrics.}
    \label{fig:radar1}
\end{figure*}

Figure~\ref{fig:radar1} illustrates the capability variance of embodied visual reasoning models across physical reasoning sub-tasks.
Leading models exhibit consistently strong performance in \emph{Visual Quality} and \emph{Background Consistency} across most categories, indicating robust low-level visual generation and scene preservation.
However, pronounced performance gaps emerge in \emph{Rule Obey} and \emph{Reasoning Accuracy}, particularly for tasks involving \emph{Measurement \& Verification}, \emph{Object Assembly}, and \emph{Situational Decision Making}, highlighting persistent challenges in instruction-following and multi-step embodied reasoning.

Notably, the discrepancy between process-level metrics and result-level metrics suggests that visually plausible intermediate states do not necessarily translate into correct reasoning outcomes.
These results reveal clear areas for improvement in precise rule compliance and reasoning reliability within embodied visual reasoning tasks.

\begin{figure*}[h!]
    \centering
    \includegraphics[width=\textwidth]{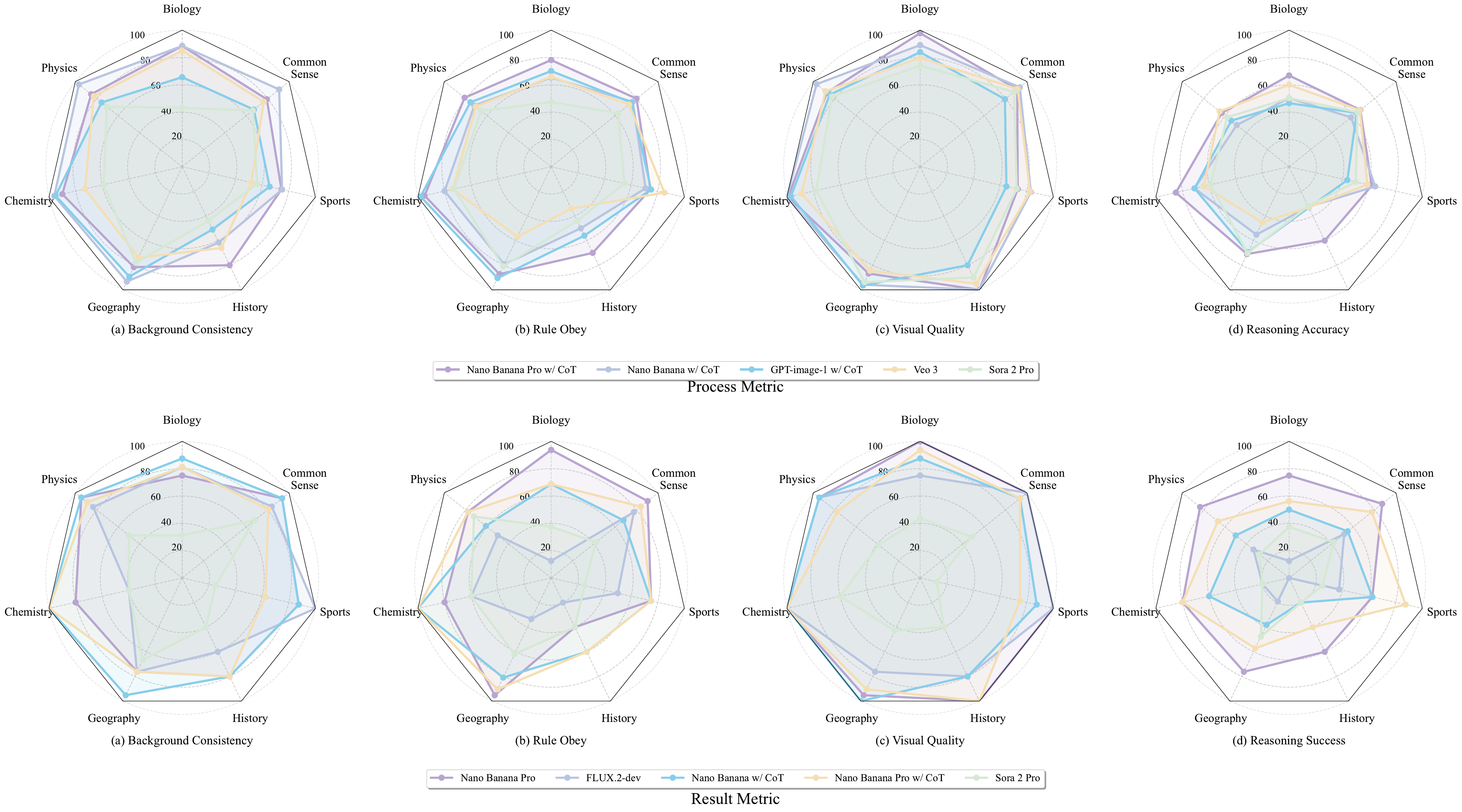}
    \caption{\textbf{Performance profiling on Knowledge reasoning tasks.} Comparison of five models across seven sub-tasks. The top and bottom rows report performance on four Process Metrics and four Result Metrics.}
    \label{fig:radar2}
\end{figure*}

Figure~\ref{fig:radar2} presents a comprehensive performance profiling of models on \emph{knowledge reasoning} tasks across seven sub-domains, including Biology, Physics, Chemistry, Geography, History, Sports, and Common Sense.
Across process-level metrics, leading models demonstrate consistently strong performance in \emph{Background Consistency} and \emph{Visual Quality} across most knowledge categories, indicating robust preservation of visual structure and semantic context.

In contrast, \emph{Rule Obey} exhibits noticeably lower and more variable performance, particularly in knowledge domains that require precise factual grounding and temporal or causal reasoning, such as \emph{History}, \emph{Geography}, and \emph{Sports}.
Result-level evaluations further reveal a clear performance degradation compared to process metrics.
While several models achieve high visual quality, their \emph{Reasoning Accuracy} and \emph{Reasoning Success} remain limited across multiple knowledge sub-tasks, highlighting persistent challenges in translating visually plausible outputs into correct knowledge-grounded reasoning outcomes.

\section{Evaluation Templates}

Table~\ref{tab:World_Knowledge_Binary_Template} to Table~\ref{tab:Maze_Navigation_CoT_Template} provides the evaluation templates used across all domains in ViGoR-Bench.
Each template specifies a standardized evaluation protocol, including task context, input image ordering, reference information, and explicit evaluation criteria.
The templates are designed to ensure consistent and reproducible assessment across different reasoning types.

% As shown in Table~\ref{tab:World_Knowledge_Binary_Template}, each template clearly defines the roles of the input image, model output, and optional ground-truth image, and focuses on instruction-following visual reasoning.
% Depending on the task, evaluation is conducted using either binary or graded criteria, covering aspects such as background consistency, rule obedience, and answer correctness.
% Together, these templates serve as unified interfaces for evaluating diverse visual reasoning tasks under a common framework.

%%%%%%%%%%%%%%%%%%%%%%%%%%%%
% Knowledge Reasoning Binary Prompt
\begin{table*}[t]
\centering
\small

\begin{tcolorbox}[
  enhanced,
  unbreakable,
  colback=gray!5,
  colframe=green!50!black,
  title=Knowledge Reasoning Binary Template
]

You are a visual reasoning evaluator specialized in world knowledge tasks.

\textbf{Task Context}\\
The model is required to solve a problem based on an input image and a given instruction.

\textbf{Input Information Structure}\\
You will receive images in the following order:
\begin{enumerate}[itemsep=0pt, topsep=0pt]
  \item \textbf{Input Image (First given image)}: The original problem image that contains the question or scenario to be solved.
  \item \textbf{Model Output (Second given image)}: The image generated by the model, showing how the model edited the input image to provide its answer. This is the primary object of evaluation.
  \item \textbf{Ground Truth Image (Third given image, if provided)}: The reference image showing the correct answer. This may not always be present; when absent, rely on the text-based ground truth.
\end{enumerate}

\textbf{Reference Information}
\begin{itemize}[itemsep=0pt, topsep=0pt]
  \item \textbf{Task Instruction}: \texttt{\{instruction\}}
  \item \textbf{Ground Truth (Standard Answer)}: \texttt{\{gt\_prompt\}}
  \item \textbf{Ground Truth Image}: Available as the third image if provided.
\end{itemize}

\textbf{Evaluation Goals:}\\
Assess the model’s output image along four binary dimensions:
\begin{enumerate}[itemsep=0pt, topsep=0pt]
  \item \textbf{Background Consistency (0 or 1)}\\
  $\rightarrow$ Does the model output retain the main structure of the input image?\\
  (Parts of the image related to the answer can be inconsistent.)

  \item \textbf{Rule Obey (0 or 1)}\\
  $\rightarrow$ Is the image edited according to the instruction?\\
  (Only the elements specified by the instruction should be changed.)

  \item \textbf{Reasoning Accuracy (0 or 1)}\\
  $\rightarrow$ Is the image edited according to the standard answer or the ground truth image(if has)?\\
  (If not completely correct, it should be considered incorrect.)

  \item \textbf{Visual Quality (0 or 1)}\\
  $\rightarrow$ Does the model output maintain high visual quality without artifacts or degradation?\\
  (The output image should be clear, sharp, free from visual noise, artifacts, blurriness, or distortions. Image content should be easily recognizable and distinguishable. Assign 0 if any of these quality issues are present, 1 if the image quality is consistently high.)
\end{enumerate}

\textbf{Output Format:}\\
Return your assessment in the following strict JSON format:
\begin{quote}\ttfamily
\{
\{
 "Background\_Consistency": 0 or 1,\\
 "Rule\_Obey": 0 or 1,\\
 "Reasoning\_Accuracy": 0 or 1,\\
 "Visual\_Quality": 0 or 1,\\
 "Explanation": \{\\
 \ \ "Background": "Explain why the background is consistent or not.",\\
 \ \ "Rule": "Explain how the rules were obeyed or violated.",\\
 \ \ "Accuracy": "Explain whether the final path reached the correct goal.",\\
 \ \ "Visual\_Quality": "Describe the overall visual quality of the output image (clarity, sharpness, absence of artifacts/noise/blur)."\}\\
\}
\}
\end{quote}

\textbf{Evaluation Principles:}
\begin{itemize}[itemsep=0pt, topsep=0pt]
  \item Assign ``1'' only if the condition is \textbf{fully satisfied} with no errors.
  \item Base your reasoning on visual comparison between input, GT, and model output.
  \item Be concise but precise in explanations.
\end{itemize}

\end{tcolorbox}
\caption{Knowledge Reasoning Binary Template.}
\vspace{-0.4cm}
\label{tab:World_Knowledge_Binary_Template}
\end{table*}

%%%%%%%%%%%%%%%%%%%%%%%%%%%%
% Knowledge Reasoning CoT Prompt
\begin{table*}[t]
\centering
\small

\begin{tcolorbox}[
  enhanced,
  unbreakable,
  colback=gray!5,
  colframe=green!50!black,
  title=Knowledge Reasoning CoT Template,
  % lower separated=false,    % 方案 A：取消上下部分的逻辑分离
  % segmentation style={draw=none} % 方案 B：直接强制不画这条线
]

You are a visual reasoning evaluator specialized in world knowledge tasks with step-by-step reasoning processes.

\textbf{Task Description}\\
The model is required to solve a problem based on an input image and a given instruction. The model generates a sequence of frames (or a video) showing its reasoning process as it progressively edits the image to provide the answer.

\textbf{Input Information Structure}\\
You will receive images/video in the following order:
\begin{enumerate}[itemsep=0pt, topsep=0pt]
  \item \textbf{First Image (Input Image)}: The original problem image that contains the question or scenario to be solved.

  \item \textbf{Middle Images/Frames (Model Output - PRIMARY EVALUATION TARGET)}: A sequence of frames or video showing the model's step-by-step reasoning process as it edits the image to provide its answer. \textbf{This is what you should evaluate.}
  \begin{itemize}[itemsep=0pt, topsep=0pt]
    \item \textbf{Important}: ALL images between the first image and the last ground truth image are part of the model's output.
    \item If a ground truth image is provided, it will appear AFTER all model frames as the very last image.
    \item If NO ground truth image is provided, ALL images after the first image are model output frames.
  \end{itemize}

  \item \textbf{Last Image (Ground Truth Reference - OPTIONAL)}: If provided, this will be the very last image in the sequence, appearing AFTER all model output frames. The reference image showing the correct answer. This is provided for comparison reference only, NOT for evaluation.
  \begin{itemize}[itemsep=0pt, topsep=0pt]
    \item \textbf{Note}: When no ground truth image is provided, rely on the text-based ground truth below.
    \item \textbf{How to identify}: The ground truth (if present) typically shows a complete final answer, distinct from the model's progressive intermediate steps.
  \end{itemize}
\end{enumerate}

\textbf{CRITICAL}: Evaluate ONLY the model's process frames (middle images/video), NOT the input or the optional ground truth reference image.

\textbf{Reference Information}
\begin{itemize}[itemsep=0pt, topsep=0pt]
  \item \textbf{Task Instruction}: \texttt{\{instruction\}}
  \item \textbf{Ground Truth (Standard Answer)}: \texttt{\{gt\_prompt\}}
  \item \textbf{Ground Truth Image}: Available as the last image if provided.
\end{itemize}

\textbf{Evaluation Dimensions}\\
Assess the model's reasoning process (middle frames) along three percentage-based dimensions (0--100):

\textbf{Background Consistency (0--100)}: In what percentage of the model's frames does the main structure of the input image remain preserved? (Parts of the image unrelated to the answer should remain unchanged; only elements directly related to solving the problem should be modified.)

\textbf{Rule Obey (0--100)}: In what percentage of the model's frames are edits made according to the instruction? (Only the elements specified by the instruction should be changed; unrelated modifications reduce this score.)

\textbf{Beneficial Action (0--100)}: In what percentage of the model's frames do the edits effectively progress toward the correct answer as specified by the ground truth? (Steps that correctly modify the image toward the GT answer increase this score; incorrect or regressive changes reduce it.)

\textbf{Visual Quality (0--100)}: In what percentage of the model's frames is the visual quality maintained at a high standard? (Frames should be clear, sharp, free from visual noise, artifacts, blurriness, or distortions. Image content should be easily recognizable and distinguishable. Frames with quality degradation, noise, artifacts, blur, or unclear content reduce this score.)

\textbf{Output Format}\\
Return your evaluation in the following JSON format:
\begin{quote}\ttfamily
\{
\{
 "Background\_Consistency": integer 0--100,\\
 "Rule\_Obey": integer 0--100,\\
 "Beneficial\_Action": integer 0--100,\\
 "Visual\_Quality": integer 0--100,\\
 "Explanation": \{\\
 \ \ "Background": "Explain how consistently the main image structure was preserved across the model's frames.",\\
 \ \ "Rule": "Explain how often the model's edits followed the instruction requirements.",\\
 \ \ "Beneficial": "Describe whether the model's progressive edits generally moved closer to the correct answer.",\\
 \ \ "Visual\_Quality": "Describe the overall visual quality across frames (clarity, sharpness, absence of artifacts/noise/blur)."\} \}\}
\end{quote}

\textbf{Evaluation Principles}
\begin{itemize}[itemsep=0pt, topsep=0pt]
  \item Percentages should be integers (e.g., 85 for 85\%).
  \item Evaluate the full temporal sequence of the \textbf{MODEL'S OUTPUT} — consider overall trends, not single-frame anomalies.
  \item A high Beneficial Action score requires consistent, correct progression toward the GT answer.
  \item Base your reasoning on visual comparison between Input, GT, and the model's process frames.
  \item Be concise but precise in explanations.
\end{itemize}

\end{tcolorbox}

\caption{Knowledge Reasoning CoT Template.}
\vspace{-0.4cm}
\label{tab:World_Knowledge_CoT_Template}
\end{table*}

%%%%%%%%%%%%%%%%%%%%%%%%%%%%
% Physical Reasoning CoT Prompt
\begin{table*}[t]
\centering
\small

\begin{tcolorbox}[
  enhanced,
  unbreakable,
  colback=gray!5,
  colframe=green!50!black,
  title=Physical Reasoning CoT Template
]

You are an expert visual reasoning evaluator for embodied action tasks with step-by-step execution processes.

\textbf{Task Description}\\
The model is required to perform physical interactions or manipulations in a real-world or simulated environment based on a given instruction. The model generates a sequence of frames (or a video) showing its reasoning and execution process as it interacts with objects, organizes items, selects tools, or completes spatial tasks.

\textbf{Input Information Structure}\\
You will receive images/video in the following order:
\begin{enumerate}[itemsep=0pt, topsep=0pt]
  \item \textbf{First Image (Input Image)}: The initial scene showing the environment, objects, and the state before any action is taken.

  \item \textbf{Middle Images/Frames (Model Output - PRIMARY EVALUATION TARGET)}: A sequence of frames or video showing the model's step-by-step action execution process (moving objects, organizing items, selecting tools, manipulating the environment). \textbf{This is what you should evaluate.}

  \item \textbf{Last Image (Ground Truth Reference - OPTIONAL)}: If provided, this will be the very last image in the sequence, appearing AFTER all model output frames. The reference image showing the correct final state. This is provided for comparison reference only, NOT for evaluation.
  \begin{itemize}[itemsep=0pt, topsep=0pt]
    \item \textbf{Note}: When no ground truth image is provided, rely on the text-based ground truth below.
  \end{itemize}
\end{enumerate}

\textbf{CRITICAL}: Evaluate ONLY the model's process frames (middle images/video), NOT the input or the optional ground truth reference image.

\textbf{Reference Information}
\begin{itemize}[itemsep=0pt, topsep=0pt]
  \item \textbf{Task Instruction}: \texttt{\{instruction\}}
  \item \textbf{Ground Truth (Text Description of Expected Outcome)}: \texttt{\{gt\_prompt\}}
  \item \textbf{Ground Truth Image}: Available as the last image if provided.
\end{itemize}

\textbf{Evaluation Dimensions}\\
Assess the model's action execution process (middle frames) along three percentage-based dimensions (0--100):

\textbf{Background Consistency (0--100)}: In what percentage of the model's frames does the environment and unrelated objects remain stable and unchanged? (Objects not involved in the task should stay in place; only objects directly mentioned in the instruction should be moved or modified.)

\textbf{Rule Obey (0--100)}: In what percentage of the model's frames are actions performed according to the instruction requirements? (Actions should follow task-specific rules such as sorting criteria, placement rules, safety constraints, or operational procedures specified in the instruction.)

\textbf{Beneficial Action (0--100)}: In what percentage of the model's frames do the actions effectively progress toward the correct final state described in the ground truth? (Steps that correctly move, organize, select, or manipulate objects toward the GT description increase this score; irrelevant, incorrect, or regressive actions reduce it.)

\textbf{Visual Quality (0--100)}: In what percentage of the model's frames is the visual quality maintained at a high standard? (Frames should be clear, sharp, free from visual noise, artifacts, blurriness, or distortions. Image content should be easily recognizable and distinguishable. Frames with quality degradation, noise, artifacts, blur, or unclear content reduce this score.)

\textbf{Output Format}\\
Return your evaluation in the following JSON format:
\begin{quote}\ttfamily
\setlength{\baselineskip}{0.9\baselineskip}
\{
\{
 "Background\_Consistency": integer 0--100,\\
 "Rule\_Obey": integer 0--100,\\
 "Beneficial\_Action": integer 0--100,\\
 "Visual\_Quality": integer 0--100,\\
 "Explanation": \{\\
 \ \ "Background": "Explain how consistently the environment and uninvolved objects remained stable across the model's frames.",\\
 \ \ "Rule": "Explain how often the model's actions followed the instruction's requirements and task-specific rules.",\\
 \ \ "Beneficial": "Whether the model's progressive actions generally moved closer to achieving the correct final state described in the ground truth text.",\\
 \ \ "Visual\_Quality": "Describe the overall visual quality across frames (clarity, sharpness, absence of artifacts/noise/blur)."\} \}
\}
\end{quote}

\textbf{Evaluation Principles}
\begin{itemize}[itemsep=0pt, topsep=0pt]
  \item Percentages should be integers (e.g., 85 for 85\%).
  \item Evaluate the full sequence of the \textbf{MODEL'S OUTPUT} — consider overall action patterns, not single-frame anomalies.
  \item A high Beneficial Action score requires consistent, goal-directed actions that align with the ground truth description.
  \item Base your reasoning on visual observation of the model's actions and comparison with the GT text description.
  \item Be concise but precise in explanations.
  \item \textbf{Remember}: Ground truth is \textbf{TEXT ONLY}.
\end{itemize}

\end{tcolorbox}

\caption{Physical Reasoning CoT Template.}
\vspace{-0.4cm}
\label{tab:Embodied_Action_CoT_Template}
\end{table*}

%%%%%%%%%%%%%%%%%%%%%%%%%%%%
% Physical Reasoning Binary Prompt
\begin{table*}[t]
\centering
\small

\begin{tcolorbox}[
  enhanced,
  unbreakable,
  colback=gray!5,
  colframe=green!50!black,
  title=Physical Reasoning Binary Template
]

You are a visual reasoning evaluator specialized in embodied action and physical manipulation tasks.

\textbf{Task Description}\\
The goal of embodied action tasks is to perform physical interactions or manipulations in a real-world or simulated environment according to a given instruction. Tasks may include sorting and organizing objects, selecting appropriate items or tools, placing objects in correct positions, adjusting settings, or completing spatial arrangements.

\textbf{Input Information Structure}\\
You will receive images in the following order:
\begin{enumerate}[itemsep=0pt, topsep=0pt]
  \item \textbf{Input Image (First given image)}: The initial scene showing the environment and objects before any action is taken.

  \item \textbf{Model Output (Second given image - PRIMARY EVALUATION TARGET)}: The final scene generated by the model showing the result of its actions. \textbf{This is what you should evaluate.}

  \item \textbf{Ground Truth Image (Third given image, if provided)}: The correct final state image. This is provided for comparison reference only, \textbf{NOT for evaluation}.
\end{enumerate}

\textbf{CRITICAL}: Evaluate ONLY the model's output image (second image), NOT the input or ground truth.

\textbf{Reference Information}
\begin{itemize}[itemsep=0pt, topsep=0pt]
  \item \textbf{Task Instruction}: \texttt{\{instruction\}}
  \item \textbf{Ground Truth (Text Description of Expected Outcome)}: \texttt{\{gt\_prompt\}}
  \item \textbf{Ground Truth Image}: Available as the third image if provided.
\end{itemize}

\textbf{Evaluation Dimensions}\\
Evaluate the model's output along three binary dimensions (0 or 1):

\textbf{Background Consistency (0 or 1)}: Does the model preserve the environment and objects not involved in the task? The overall scene layout, uninvolved objects' positions, and environmental elements should remain unchanged unless explicitly required by the instruction.

\textbf{Rule Obey (0 or 1)}: Does the model's final result follow all the rules and requirements specified in the instruction? This includes sorting criteria (e.g., by color, size, type), placement rules (e.g., specific locations, correct containers), selection criteria (e.g., choosing appropriate tools or items), safety constraints, operational procedures, or any other task-specific requirements.

\textbf{Reasoning Accuracy (0 or 1)}: Does the model's final result match the ground truth? All objects should be in their correct final positions, all required actions should be completed, and the final state should align with what is described in the GT text.

\textbf{Visual Quality (0 or 1)}: Does the model output maintain high visual quality without artifacts or degradation? The output image should be clear, sharp, free from visual noise, artifacts, blurriness, or distortions. Image content should be easily recognizable and distinguishable. Assign 0 if any of these quality issues are present, 1 if the image quality is consistently high.

\textbf{Output Format}\\
Provide your evaluation strictly in JSON format:
\begin{quote}\ttfamily
\{
\{
 "Background\_Consistency": 0 or 1,\\
 "Rule\_Obey": 0 or 1,\\
 "Reasoning\_Accuracy": 0 or 1,\\
 "Visual\_Quality": 0 or 1,\\
 "Explanation": \{"Background": "Describe whether the environment and uninvolved objects were preserved correctly.",\\
 \ \ "Rule": "Describe whether all instruction requirements and task-specific rules were followed.",\\
 \ \ "Accuracy": "Describe whether the final result matches the ground truth text description.",\\
 \ \ "Visual\_Quality": "Describe the overall visual quality of the output image (clarity, sharpness, absence of artifacts/noise/blur)."\} \}\}
\end{quote}

\textbf{Evaluation Principles}
\begin{itemize}[itemsep=0pt, topsep=0pt]
  \item Assign ``1'' only if the criterion is fully satisfied without errors.
  \item Evaluate based on visual observation of the final result and comparison with the GT text description.
  \item Partial correctness (e.g., some items correctly placed but others wrong) counts as 0 for that dimension.
  \item Keep explanations factual, objective, and concise.
  \item \textbf{Remember}: Ground truth is \textbf{TEXT ONLY} — assess whether the visual output matches the textual description.
\end{itemize}

\end{tcolorbox}

\caption{Physical Reasoning Binary Template.}
\vspace{-0.4cm}
\label{tab:Embodied_Action_Binary_Template}
\end{table*}

%%%%%%%%%%%%%%%%%%%%%%%%%%%%
% Sudoku CoT Prompt
\begin{table*}[t]
\centering
\small

\begin{tcolorbox}[
  enhanced,
  unbreakable,
  colback=gray!5,
  colframe=green!50!black,
  title=Symbolic Reasoning - Sudoku CoT Template
]

You are an expert visual reasoning evaluator for step-by-step Sudoku-solving processes.

\textbf{Task Description}\\
The model gradually fills in the Sudoku grid to reach a complete solution according to standard Sudoku rules. Each frame in the process may represent a partial filling stage or a correction step.

\textbf{Input Information Structure}\\
You will receive images or video in the following order:
\begin{enumerate}[itemsep=0pt, topsep=0pt]
  \item \textbf{First Image (Input Image)}: The initial Sudoku puzzle with fixed given digits (typically shown in blue).

  \item \textbf{Middle Images or Frames (Model Output - PRIMARY EVALUATION TARGET)}: A sequence of frames or a video showing the model's step-by-step Sudoku solving process (gradually filling in digits). \textbf{This is what you should evaluate.}

  \item \textbf{Last Image (Ground Truth Reference)}: The correct final Sudoku solution showing all digits filled correctly. This is provided for comparison reference only, \textbf{NOT for evaluation}.
\end{enumerate}

\textbf{CRITICAL}: Evaluate ONLY the model's process frames (middle images/video), NOT the input or ground truth reference image.

\textbf{Reference Information}
\begin{itemize}[itemsep=0pt, topsep=0pt]
  \item \textbf{Instruction}: ``Fill the Sudoku grid according to standard Sudoku rules so that each row, column, and subgrid contains all digits exactly once.''
  \item \textbf{Ground Truth}: The last image shows the correct final Sudoku solution for comparison.
\end{itemize}

\textbf{Evaluation Dimensions}\\
Evaluate the full reasoning process along three percentage-based dimensions (0--100):

\textbf{Background Consistency (0--100)}: In what percentage of frames are the grid structure and original given digits preserved correctly? Altered blue digits, misplaced lines, or corrupted layout reduce this score.

\textbf{Rule Obey (0--100)}: In what percentage of frames does the partial filling obey Sudoku rules? Each intermediate grid should not violate uniqueness constraints within rows, columns, or subgrids.

\textbf{Beneficial Action (0--100)}: In what percentage of frames do the filled or corrected digits move the grid closer to the GT solution? Correctly placed digits or valid logical progressions increase this score; random or regressive changes reduce it.

\textbf{Visual Quality (0--100)}: In what percentage of frames is the visual quality maintained at a high standard? (Frames should be clear, sharp, free from visual noise, artifacts, blurriness, or distortions. Image content should be easily recognizable and distinguishable. Frames with quality degradation, noise, artifacts, blur, or unclear content reduce this score.)

\textbf{Output Format}\\
Provide your evaluation strictly in JSON format:
\begin{quote}\ttfamily
\{
\{
 "Background\_Consistency": integer 0--100,\\
 "Rule\_Obey": integer 0--100,\\
 "Beneficial\_Action": integer 0--100,\\
 "Visual\_Quality": integer 0--100,\\
 "Explanation": \{\\
 \ \ "Background": "Explain how stable the grid and given digits remained during solving.",\\
 \ \ "Rule": "Explain how frequently intermediate states obeyed Sudoku rules.",\\
 \ \ "Beneficial": "Explain how much of the model's step-by-step reasoning effectively advanced toward the correct solution.",\\
 \ \ "Visual\_Quality": "Describe the overall visual quality across frames (clarity, sharpness, absence of artifacts/noise/blur)."\\
 \}\\
\}
\}
\end{quote}

\textbf{Evaluation Principles}
\begin{itemize}[itemsep=0pt, topsep=0pt]
  \item Percentages must be integers (e.g., 90 for 90\%).
  \item Assess the full temporal reasoning sequence, not just the final frames.
  \item High ``Rule Obey'' requires most intermediate grids to be logically valid.
  \item High ``Beneficial Action'' requires consistent progress toward the GT completion.
  \item Evaluate purely based on visual and logical evidence, without inference beyond shown content.
\end{itemize}

\end{tcolorbox}

\caption{Symbolic Reasoning - Sudoku CoT Template.}
\vspace{-0.4cm}
\label{tab:Sudoku_Solver_CoT_Template}
\end{table*}

%%%%%%%%%%%%%%%%%%%%%%%%%%%%
% Symbolic Reasoning - Sudoku Binary Prompt
\begin{table*}[t]
\centering
\small

\begin{tcolorbox}[
  enhanced,
  unbreakable,
  colback=gray!5,
  colframe=green!50!black,
  title=Symbolic Reasoning - Sudoku Binary Template
]
You are a visual reasoning evaluator specializing in Sudoku-solving tasks.

\textbf{Task Description}\\
The goal of this task is to fill the Sudoku grid according to standard Sudoku rules so that each row, column, and subgrid contains all digits exactly once. The puzzle may vary in grid size (e.g., 2$\times$2, 4$\times$4, 6$\times$6, 7$\times$7). The model must produce a completed Sudoku grid that follows all structural and logical constraints.

\textbf{Input Information Structure}\\
You will receive images in the following order:
\begin{enumerate}[itemsep=0pt, topsep=0pt]
  \item \textbf{Input Image (First given image)}: The initial Sudoku puzzle with given numbers (partial grid, typically shown in blue).

  \item \textbf{Model Output (Second given image - PRIMARY EVALUATION TARGET)}: The filled Sudoku grid generated by the model. \textbf{This is what you should evaluate.}

  \item \textbf{Ground Truth Image (Third given image, Optional)}: The correct completed Sudoku solution. This is provided for comparison reference only, \textbf{NOT for evaluation}.
\end{enumerate}

\textbf{CRITICAL}: Evaluate ONLY the model's output image (second image), NOT the input or ground truth image.

\textbf{Reference Information}
\begin{itemize}[itemsep=0pt, topsep=0pt]
  \item \textbf{Instruction}: ``Fill the Sudoku grid according to standard Sudoku rules so that each row, column, and subgrid contains all digits exactly once.''
  \item \textbf{Ground Truth}: The third image shows the correct completed Sudoku solution for comparison.
\end{itemize}

\textbf{Evaluation Dimensions}\\
Evaluate the model's output along three binary dimensions (0 or 1):

\textbf{Background Consistency (0 or 1)}: Does the model preserve the grid layout, line thickness, cell structure, and fixed numbers (the given blue digits) without modification or deletion?

\textbf{Rule Obey (0 or 1)}: Does the filled Sudoku strictly follow standard rules? Each row, column, and subgrid must contain all unique digits without repetition.

\textbf{Reasoning Accuracy (0 or 1)}: Does the final solution match the ground truth exactly? All digits must correspond to the GT positions and values.

\textbf{Visual Quality (0 or 1)}: Does the model output maintain high visual quality without artifacts or degradation? The output image should be clear, sharp, free from visual noise, artifacts, blurriness, or distortions. Image content should be easily recognizable and distinguishable. Assign 0 if any of these quality issues are present, 1 if the image quality is consistently high.

\textbf{Output Format}\\
Provide your evaluation strictly in JSON format:
\begin{quote}\ttfamily
\{
\{
 "Background\_Consistency": 0 or 1,\\
 "Rule\_Obey": 0 or 1,\\
 "Reasoning\_Accuracy": 0 or 1,\\
 "Visual\_Quality": 0 or 1,\\
 "Explanation": \{\\
 \ \ "Background": "Describe whether the grid structure and given digits were preserved.",\\
 \ \ "Rule": "Describe whether all rows, columns, and subgrids obey Sudoku uniqueness rules.",\\
 \ \ "Accuracy": "Describe whether the filled digits match the GT solution exactly.",\\
 \ \ "Visual\_Quality": "Describe the overall visual quality of the output image (clarity, sharpness, absence of artifacts/noise/blur)."\\
 \}\\
\}
\}
\end{quote}

\textbf{Evaluation Principles}
\begin{itemize}[itemsep=0pt, topsep=0pt]
  \item Assign ``1'' only if the criterion is fully satisfied.
  \item Evaluate based on visual inspection and logical Sudoku validity.
  \item Partial correctness (e.g., one subgrid error) counts as 0 for that dimension.
  \item Keep explanations factual and concise.
\end{itemize}

\end{tcolorbox}

\caption{Symbolic Reasoning - Sudoku Binary Template.}
\vspace{-0.4cm}
\label{tab:Symbolic Reasoning - Sudoku_Solver_Binary_Template}
\end{table*}

%%%%%%%%%%%%%%%%%%%%%%%%%%%%
% Jigsaw Puzzle CoT Prompt
\begin{table*}[t]
\centering
\small

\begin{tcolorbox}[
  enhanced,
  unbreakable,
  colback=gray!5,
  colframe=green!50!black,
  title=Symbolic Reasoning - Jigsaw Puzzle CoT Template
]

You are an expert visual reasoning evaluator for jigsaw puzzle restoration processes.

\textbf{Task Description}\\
The task involves progressively restoring a scrambled jigsaw puzzle to its original image. The model generates a sequence of frames (or a video) showing its reasoning and assembly process, where puzzle tiles are gradually moved, rotated, or placed in position.

\textbf{Input Information Structure}\\
You will receive images or video in the following order:
\begin{enumerate}[itemsep=0pt, topsep=0pt]
  \item \textbf{First Image (Input Image)}: The scrambled jigsaw puzzle image showing tiles in incorrect positions.

  \item \textbf{Middle Images or Frames (Model Output - PRIMARY EVALUATION TARGET)}: A sequence of frames or a video showing the model's step-by-step puzzle restoration process (moving, rotating, or placing tiles). \textbf{This is what you should evaluate.}

  \item \textbf{Last Image (Ground Truth Reference)}: The correct complete image showing the original unscrambled layout. This is provided for comparison reference only, \textbf{NOT for evaluation}.
\end{enumerate}

\textbf{CRITICAL}: Evaluate ONLY the model's process frames (middle images/video), NOT the input or ground truth.

\textbf{Reference Information}
\begin{itemize}[itemsep=0pt, topsep=0pt]
  \item \textbf{Instruction}: ``Restore the scrambled jigsaw puzzle to its original complete image.''
  \item \textbf{Ground Truth}: The last image shows the correct complete layout for comparison.
\end{itemize}

\textbf{Evaluation Dimensions}\\
Evaluate the overall reasoning process along three percentage-based dimensions (0--100):

\textbf{Background Consistency (0--100)}: In what percentage of frames does the global color, lighting, and visual continuity remain stable? Frames with visual noise, artifact introduction, or tile color mismatch reduce the score.

\textbf{Rule Obey (0--100)}: In what percentage of frames does the model follow jigsaw reconstruction rules? Each frame should show valid moves (placing or rotating tiles without overlaps or deletions).

\textbf{Beneficial Action (0--100)}: In what percentage of frames do the actions effectively move the puzzle toward completion? Frames that position or correct tiles to their GT location are beneficial; random or regressive moves reduce the score.

\textbf{Visual Quality (0--100)}: In what percentage of frames is the visual quality maintained at a high standard? Frames should be clear, sharp, free from visual noise, artifacts, blurriness, or distortions. Image content should be easily recognizable and distinguishable. Frames with quality degradation, noise, artifacts, blur, or unclear content reduce this score.

\textbf{Output Format}\\
Provide your evaluation strictly in JSON format:
\begin{quote}\ttfamily
\{
\{
 "Background\_Consistency": integer 0--100,\\
 "Rule\_Obey": integer 0--100,\\
 "Beneficial\_Action": integer 0--100,\\
 "Visual\_Quality": integer 0--100,\\
 "Explanation": \{\\
 \ \ "Background": "Explain how stable and visually consistent the background remained.",\\
 \ \ "Rule": "Explain how often the puzzle assembly rules were respected (no overlaps, valid placements).",\\
 \ \ "Beneficial": "Explain how frequently the model made progress toward the correct GT arrangement.",\\
 \ \ "Visual\_Quality": "Describe the overall visual quality across frames (clarity, sharpness, absence of artifacts/noise/blur)."\\
 \}\\
\}
\}
\end{quote}

\textbf{Evaluation Principles}
\begin{itemize}[itemsep=0pt, topsep=0pt]
  \item Percentages must be integers (e.g., 90 for 90\%).
  \item Evaluate the full sequence, not only the last few frames.
  \item High scores require consistent, rule-abiding, and progressive assembly behavior.
  \item Visual comparison with the ground truth should guide all judgments.
\end{itemize}

\end{tcolorbox}

\caption{Symbolic Reasoning - Jigsaw Puzzle CoT Template.}
\vspace{-0.4cm}
\label{tab:Symbolic Reasoning - Jigsaw_Restoration_CoT_Template}
\end{table*}

%%%%%%%%%%%%%%%%%%%%%%%%%%%%
% Jigsaw Puzzle Binary Prompt
\begin{table*}[t]
\centering
\small

\begin{tcolorbox}[
  enhanced,
  unbreakable,
  colback=gray!5,
  colframe=green!50!black,
  title=Symbolic Reasoning - Jigsaw Puzzle Binary Template
]

You are a visual reasoning evaluator specialized in image reconstruction and jigsaw puzzle restoration tasks.

\textbf{Task Description}\\
The goal of this task is to restore a scrambled jigsaw puzzle image back to its original complete image. The puzzle may have different grid sizes (e.g., 2$\times$2, 4$\times$4, 7$\times$7), and the model must correctly reassemble all tiles in their proper positions to match the original image.

\textbf{Input Information Structure}\\
You will receive images in the following order:
\begin{enumerate}[itemsep=0pt, topsep=0pt]
  \item \textbf{Input Image (First given image)}: The scrambled jigsaw puzzle image showing tiles in incorrect positions.

  \item \textbf{Model Output (Second given image - PRIMARY EVALUATION TARGET)}: The image produced by the model showing its reconstructed result. \textbf{This is what you should evaluate.}

  \item \textbf{Ground Truth Image (Third given image, if provided)}: The correct full image showing the original unshuffled arrangement. This is provided for comparison reference only, \textbf{NOT for evaluation}.
\end{enumerate}

\textbf{CRITICAL}: Evaluate ONLY the model's output image (second image), NOT the input or ground truth.

\textbf{Reference Information}
\begin{itemize}[itemsep=0pt, topsep=0pt]
  \item \textbf{Instruction}: ``Restore the scrambled jigsaw puzzle to its original complete image.''
  \item \textbf{Ground Truth}: The third image shows the correct full image for comparison.
\end{itemize}

\textbf{Evaluation Dimensions}\\
Evaluate the model's output along three binary dimensions (0 or 1):

\textbf{Background Consistency (0 or 1)}: Are all visual elements (color tones, lighting, and texture continuity) globally consistent after restoration? The restored image should not introduce artifacts, gaps, or color mismatches between tiles.

\textbf{Rule Obey (0 or 1)}: Does the model follow the basic jigsaw reconstruction rules? Each tile should be placed once, without overlap or missing regions, and edges should align correctly.

\textbf{Reasoning Accuracy (0 or 1)}: Is the final restored image visually identical or sufficiently close to the ground truth? The global spatial arrangement must fully match the GT layout.

\textbf{Visual Quality (0 or 1)}: Does the model output maintain high visual quality without artifacts or degradation? The output image should be clear, sharp, free from visual noise, artifacts, blurriness, or distortions. Image content should be easily recognizable and distinguishable. Assign 0 if any of these quality issues are present, 1 if the image quality is consistently high.

\textbf{Output Format}\\
Provide your evaluation strictly in JSON format:
\begin{quote}\ttfamily
\{
\{
 "Background\_Consistency": 0 or 1,\\
 "Rule\_Obey": 0 or 1,\\
 "Reasoning\_Accuracy": 0 or 1,\\
 "Visual\_Quality": 0 or 1,\\
 "Explanation": \{\\
 \ \ "Background": "Describe whether color, lighting, and texture continuity were preserved.",\\
 \ \ "Rule": "Describe if all tiles were used and correctly aligned without overlaps or missing areas.",\\
 \ \ "Accuracy": "Describe whether the final image matches the GT layout and visual composition.",\\
 \ \ "Visual\_Quality": "Describe the overall visual quality of the output image (clarity, sharpness, absence of artifacts/noise/blur)."\\
 \}\\
\}
\}
\end{quote}

\textbf{Evaluation Principles}
\begin{itemize}[itemsep=0pt, topsep=0pt]
  \item Assign ``1'' only if the condition is fully satisfied.
  \item Evaluate based on visual alignment and texture continuity between Input, GT, and Model Output.
  \item Minor seam visibility does not affect correctness if the arrangement is fully correct.
  \item Explanations should be objective and concise.
\end{itemize}

\end{tcolorbox}

\caption{Symbolic Reasoning - Jigsaw Puzzle Binary Template.}
\vspace{-0.4cm}
\label{tab:Symbolic Reasoning - Jigsaw_Restoration_Binary_Template}
\end{table*}

%%%%%%%%%%%%%%%%%%%%%%%%%%%%
% Function Plotting CoT Prompt
\begin{table*}[t]
\centering
\small

\begin{tcolorbox}[
  enhanced,
  unbreakable,
  colback=gray!5,
  colframe=green!50!black,
  title=Symbolic Reasoning - Function Plotting CoT Template
]

You are an expert visual reasoning evaluator for function plotting processes.

\textbf{Task Description}\\
The model generates a sequence of frames (or a video) that illustrates how a mathematical function $y=f(x)$ is drawn step by step over a specified domain. Each frame may show intermediate elements such as coordinate axes, sample points, partial curves, or annotations.

\textbf{Input Information Structure}\\
You will receive images or video in the following order:
\begin{enumerate}[itemsep=0pt, topsep=0pt]
  \item \textbf{First Image (Input Image)}: The original image showing the mathematical formula and domain specification (e.g., $y = x^2 \cdot \sin^2(x) + x$ over $x \in [-10.0, 10.0]$).

  \item \textbf{Middle Images or Frames (Model Output - PRIMARY EVALUATION TARGET)}: A sequence of frames or a video showing the model's step-by-step function plotting process (coordinate axes setup, sample points, partial curves). \textbf{This is what you should evaluate.}

  \item \textbf{Last Image (Ground Truth Reference)}: The correct completed curve plot and, if applicable, an idealized plotting progression. This is provided for comparison reference only, \textbf{NOT for evaluation}.
\end{enumerate}

\textbf{CRITICAL}: Evaluate ONLY the model's process frames (middle images/video), NOT the input or ground truth.

\textbf{Reference Information}
\begin{itemize}[itemsep=0pt, topsep=0pt]
  \item \textbf{Instruction}: ``Draw the curve corresponding to the mathematical expression shown in the image over the specified domain.''
  \item \textbf{Ground Truth}: The last image shows the correct completed curve plot for comparison.
\end{itemize}

\textbf{Evaluation Dimensions}\\
Evaluate the overall reasoning process along the following percentage-based dimensions (0--100):

\textbf{Background Consistency (0--100)}: In what percentage of frames does the formula text, domain, and coordinate system remain stable and correct? Alterations to labels or axis structure lower the score.

\textbf{Rule Obey (0--100)}: In what percentage of frames does the progressive curve follow the correct mathematical rule? Points or partial lines must be located where $f(x)$ predicts, and the specified domain range must not be exceeded.

\textbf{Beneficial Action (0--100)}: In what percentage of frames do the drawing actions meaningfully progress toward completing the correct curve? Steps that add correct data points or refine the curve are counted as beneficial; random or inconsistent updates are penalized.

\textbf{Visual Quality (0--100)}: In what percentage of frames is the visual quality maintained at a high standard? Frames should be clear, sharp, free from visual noise, artifacts, blurriness, or distortions. Image content should be easily recognizable and distinguishable.

\textbf{Output Format}\\
Return your evaluation in the following strict JSON format:
\begin{quote}\ttfamily
\{
\{
 "Background\_Consistency": integer 0--100,\\
 "Rule\_Obey": integer 0--100,\\
 "Beneficial\_Action": integer 0--100,\\
 "Visual\_Quality": integer 0--100,\\
 "Explanation": \{\\
 \ \ "Background": "Explain whether the formula and coordinate system remained consistent.",\\
 \ \ "Rule": "Describe how often the intermediate steps followed the correct mathematical behavior.",\\
 \ \ "Beneficial": "Describe how effectively each step contributed to completing the correct curve.",\\
 \ \ "Visual\_Quality": "Describe the overall visual quality across frames (clarity, sharpness, absence of artifacts/noise/blur)."\} \}
\}
\end{quote}

\textbf{Evaluation Principles}
\begin{itemize}[itemsep=0pt, topsep=0pt]
  \item Percentages must be integers (e.g., 90 for 90\%).
  \item Evaluate the full sequence of frames, focusing on stability and progressive correctness.
  \item High Rule Obey requires most intermediate plots to conform to the true mathematical function.
  \item High Beneficial Action requires consistent forward progress toward the ground truth curve.
  \item Base your evaluation solely on visual and mathematical reasoning; do not infer beyond the given content.
\end{itemize}

\end{tcolorbox}

\caption{Symbolic Reasoning - Function Plotting CoT Template.}
\vspace{-0.4cm}
\label{tab:Symbolic Reasoning - Curve_Drawing_CoT_Template}
\end{table*}

%%%%%%%%%%%%%%%%%%%%%%%%%%%%
% Function Plotting Binary Prompt
\begin{table*}[t]
\centering
\small

\begin{tcolorbox}[
  enhanced,
  unbreakable,
  colback=gray!5,
  colframe=green!50!black,
  title=Symbolic Reasoning - Function Plotting Binary Template
]

You are a visual reasoning evaluator specialized in mathematical visualization and function plotting.

\textbf{Task Description}\\
The task is to draw the curve corresponding to a given mathematical expression over a specified domain. The model receives:
\begin{itemize}[itemsep=0pt, topsep=0pt]
  \item A mathematical formula (e.g., $y = x^2 \cdot \sin^2(x) + x$)
  \item A domain (e.g., $x \in [-10.0, 10.0]$)
\end{itemize}
and is instructed to plot the correct function curve across this domain.

\textbf{Input Information Structure}\\
You will receive images in the following order:
\begin{enumerate}[itemsep=0pt, topsep=0pt]
  \item \textbf{Input Image (First given image)}: The image showing the mathematical formula and the domain specification.

  \item \textbf{Model Output (Second given image - PRIMARY EVALUATION TARGET)}: The image generated by the model showing its plotted curve. \textbf{This is what you should evaluate.}

  \item \textbf{Ground Truth Image (Third given image, Optional)}: The correct plot of the function over the same domain. This is provided for comparison reference only, \textbf{NOT for evaluation}.
\end{enumerate}

\textbf{CRITICAL}: Evaluate ONLY the model's output image (second image), NOT the input or ground truth.

\textbf{Reference Information}
\begin{itemize}[itemsep=0pt, topsep=0pt]
  \item \textbf{Instruction}: ``Draw the curve corresponding to the mathematical expression shown in the image over the specified domain.''
  \item \textbf{Ground Truth}: The third image shows the correct plot of the function for comparison.
\end{itemize}

\textbf{Evaluation Dimensions}\\
Evaluate the model's output image along the following binary dimensions (0 or 1):

\textbf{Background Consistency (0 or 1)}: Does the model preserve all non-plot elements from the input, including the formula box, domain text, axis labels, and overall layout?

\textbf{Rule Obey (0 or 1)}: Does the plotted curve obey the mathematical rule defined by the formula and the specified domain? The curve should exhibit correct functional behavior (e.g., symmetry, periodicity, growth trends) and stay within the valid $x$-range.

\textbf{Reasoning Accuracy (0 or 1)}: Is the plotted curve visually identical or sufficiently close to the ground truth plot? The overall shape, peaks, crossings, and range must align with the GT.

\textbf{Visual Quality (0 or 1)}: Does the model output maintain high visual quality without artifacts or degradation? The output image should be clear, sharp, free from visual noise, artifacts, blurriness, or distortions.

\textbf{Output Format}\\
Provide your evaluation strictly in the following JSON format:
\begin{quote}\ttfamily
\{
\{
 "Background\_Consistency": 0 or 1,\\
 "Rule\_Obey": 0 or 1,\\
 "Reasoning\_Accuracy": 0 or 1,\\
 "Visual\_Quality": 0 or 1,\\
 "Explanation": \{\\
 \ \ "Background": "Describe whether the non-plot text such as formula and domain remained consistent.",\\
 \ \ "Rule": "Describe whether the curve obeyed the given mathematical formula within the domain.",\\
 \ \ "Accuracy": "Describe whether the plotted curve matches the ground truth in shape and range.",\\
 \ \ "Visual\_Quality": "Describe the overall visual quality of the output image (clarity, sharpness, absence of artifacts/noise/blur)."\\
 \}\\
\}
\}
\end{quote}

\textbf{Evaluation Principles}
\begin{itemize}[itemsep=0pt, topsep=0pt]
  \item Assign ``1'' only if the criterion is fully satisfied without visual or logical errors.
  \item Evaluation is based on the visual and mathematical alignment between Input, Ground Truth, and Model Output.
  \item Deviations in function shape or domain range result in a score of 0 for Rule Obey and Reasoning Accuracy.
  \item Explanations should be precise, objective, and concise.
\end{itemize}

\end{tcolorbox}

\caption{Symbolic Reasoning - Function Plotting Binary Template.}
\vspace{-0.4cm}
\label{tab:Symbolic Reasoning - Curve_Drawing_Binary_Template}
\end{table*}

%%%%%%%%%%%%%%%%%%%%%%%%%%%%
% Algebraic Calculation CoT Prompt
\begin{table*}[t]
\centering
\small

\begin{tcolorbox}[
  enhanced,
  unbreakable,
  colback=gray!5,
  colframe=green!50!black,
  title=Symbolic Reasoning - Algebraic Calculation CoT Template
]

You are an expert visual reasoning evaluator for mathematical problem-solving processes, focusing on algebraic equations.

\textbf{Task Description}\\
The task involves solving algebraic equations or systems (linear or quadratic) step by step. Each frame or intermediate image shows the model's reasoning process, such as simplifying, substituting, or solving for variables.

\textbf{Input Information Structure}\\
You will receive images or video in the following order:
\begin{enumerate}[itemsep=0pt, topsep=0pt]
  \item \textbf{First Image (Input Image)}: The original problem containing the equation or equations to be solved.

  \item \textbf{Middle Images or Frames (Model Output - PRIMARY EVALUATION TARGET)}: A sequence of frames or a video showing the model's step-by-step algebraic reasoning process (simplifying, substituting, solving for variables). \textbf{This is what you should evaluate.}

  \item \textbf{Last Image (Ground Truth Reference)}: The correct final solution showing the complete solving process and final answer. This is provided for comparison reference only, \textbf{NOT for evaluation}.
\end{enumerate}

\textbf{CRITICAL}: Evaluate ONLY the model's process frames (middle images/video), NOT the input or ground truth.

\textbf{Reference Information}
\begin{itemize}[itemsep=0pt, topsep=0pt]
  \item \textbf{Instruction}: ``Please solve this equation.''
  \item \textbf{Ground Truth}: The last image shows the correct final solution and a logically valid sequence of transformations for comparison.
\end{itemize}

\textbf{Evaluation Dimensions}\\
Evaluate the reasoning process along the following percentage-based dimensions (0--100):

\textbf{Background Consistency (0--100)}: In what percentage of frames does the original equation text remain intact and readable? Changes to problem statements or unrelated additions reduce this score.

\textbf{Rule Obey (0--100)}: In what percentage of steps are algebraic operations performed correctly? Valid transformations include addition, subtraction, factoring, substitution, and simplification according to algebraic laws.

\textbf{Beneficial Action (0--100)}: In what percentage of steps does the reasoning progress toward the correct solution? Steps that correctly reduce equation complexity or move closer to the ground truth contribute positively.

\textbf{Visual Quality (0--100)}: In what percentage of frames is the visual quality maintained at a high standard? Frames should be clear, sharp, free from visual noise, artifacts, blurriness, or distortions.

\textbf{Output Format}\\
Return your evaluation strictly in the following JSON format:
\begin{quote}\ttfamily
\{
\{
 "Background\_Consistency": integer 0--100,\\
 "Rule\_Obey": integer 0--100,\\
 "Beneficial\_Action": integer 0--100,\\
 "Visual\_Quality": integer 0--100,\\
 "Explanation": \{\\
 \ \ "Background": "Explain how consistently the problem layout was preserved across steps.",\\
 \ \ "Rule": "Describe how often algebraic rules were obeyed in the reasoning process.",\\
 \ \ "Beneficial": "Describe how effectively the model's steps advanced toward the correct solution.",\\
 \ \ "Visual\_Quality": "Describe the overall visual quality across frames (clarity, sharpness, absence of artifacts/noise/blur)."\\
 \}\\
\}
\}
\end{quote}

\textbf{Evaluation Principles}
\begin{itemize}[itemsep=0pt, topsep=0pt]
  \item Percentages must be integers.
  \item Consider the full reasoning sequence, not isolated steps.
  \item Reward logical consistency and stepwise correctness even if the final result is not exact.
  \item Base all judgments on visual comparison between Input, Ground Truth, and Process frames.
  \item Explanations should be concise and factual.
\end{itemize}

\end{tcolorbox}

\caption{Symbolic Reasoning - Algebraic Calculation CoT Template.}
\vspace{-0.4cm}
\label{tab:Algebra_Solution_CoT_Template}
\end{table*}

%%%%%%%%%%%%%%%%%%%%%%%%%%%%
% Algebraic Calculation Binary Prompt
\begin{table*}[t]
\centering
\small

\begin{tcolorbox}[
  enhanced,
  unbreakable,
  colback=gray!5,
  colframe=green!50!black,
  title=Symbolic Reasoning - Algebraic Calculation Binary Template
]

You are a visual reasoning evaluator specialized in mathematical problem solving, specifically algebraic equations.

\textbf{Task Description}\\
The goal of this task is to solve an algebraic equation or system of equations shown in the image. The equations may include:
\begin{itemize}[itemsep=0pt, topsep=0pt]
  \item Single-variable linear equations (e.g., $2x + 3 = 5$)
  \item Single-variable quadratic equations (e.g., $x^2 + 2x - 3 = 0$)
  \item Two-variable linear equations (e.g., $2x + 9y = 19$ and $2x + 6y = 8$)
\end{itemize}

\textbf{Input Information Structure}\\
You will receive images in the following order:
\begin{enumerate}[itemsep=0pt, topsep=0pt]
  \item \textbf{Input Image (First given image)}: The original problem containing the equation or equations to be solved.

  \item \textbf{Model Output (Second given image - PRIMARY EVALUATION TARGET)}: The image produced by the model containing its proposed solution. \textbf{This is what you should evaluate.}

  \item \textbf{Ground Truth Image (Third given image, if provided)}: The correct final answer(s) and solution form. This is provided for comparison reference only, \textbf{NOT for evaluation}.
\end{enumerate}

\textbf{CRITICAL}: Evaluate ONLY the model's output image (second image), NOT the input or ground truth.

\textbf{Reference Information}
\begin{itemize}[itemsep=0pt, topsep=0pt]
  \item \textbf{Instruction}: ``Please solve this equation.''
  \item \textbf{Ground Truth}: The third image shows the correct final answer(s) and solution form for comparison.
\end{itemize}

\textbf{Evaluation Dimensions}\\
Evaluate the model output along the following binary dimensions (0 or 1):

\textbf{Background Consistency (0 or 1)}: Does the output preserve the original problem text and structure? The equation layout, fonts, and non-solution content should remain intact and legible.

\textbf{Rule Obey (0 or 1)}: Does the model follow correct algebraic transformation rules? This includes valid symbolic manipulation (e.g., addition, subtraction, factoring, substitution). Violations such as illegal simplification or missing logical steps result in 0.

\textbf{Reasoning Accuracy (0 or 1)}: Is the final solution correct and fully consistent with the ground truth? For single-variable equations, check the numerical or symbolic answer. For multi-variable systems, all variable pairs $(x, y)$ must match the GT values.

\textbf{Visual Quality (0 or 1)}: Does the model output maintain high visual quality without artifacts or degradation? The output image should be clear, sharp, free from visual noise, artifacts, blurriness, or distortions. Image content should be easily recognizable and distinguishable.

\textbf{Output Format}\\
Provide your evaluation strictly in the following JSON format:
\begin{quote}\ttfamily
\{
\{
 "Background\_Consistency": 0 or 1,\\
 "Rule\_Obey": 0 or 1,\\
 "Reasoning\_Accuracy": 0 or 1,\\
 "Visual\_Quality": 0 or 1,\\
 "Explanation": \{\\
 \ \ "Background": "Describe whether the equation layout was preserved.",\\
 \ \ "Rule": "Describe whether algebraic steps followed mathematical rules.",\\
 \ \ "Accuracy": "Describe whether the final result matches the correct solution.",\\
 \ \ "Visual\_Quality": "Describe the overall visual quality of the output image (clarity, sharpness, absence of artifacts/noise/blur)."\\
 \}\\
\}
\}
\end{quote}

\textbf{Evaluation Principles}
\begin{itemize}[itemsep=0pt, topsep=0pt]
  \item Assign ``1'' only if the criterion is fully satisfied without mistakes.
  \item Visual comparison should ensure no loss of context or formatting corruption.
  \item For partial correctness (e.g., right procedure but wrong result), only \textbf{Rule Obey} may receive 1; all other dimensions must be 0.
\end{itemize}

\end{tcolorbox}

\caption{Symbolic Reasoning - Algebraic Calculation Binary Template.}
\vspace{-0.4cm}
\label{tab:Algebra_Solution_Binary_Template}
\end{table*}

%%%%%%%%%%%%%%%%%%%%%%%%%%%%
% Block Building CoT Prompt
\begin{table*}[t]
\centering
\small

\begin{tcolorbox}[
  enhanced,
  unbreakable,
  colback=gray!5,
  colframe=green!50!black,
  title=Symbolic Reasoning - Block Building CoT Template
]

You are an expert visual reasoning evaluator for LEGO blocks processes.

\textbf{Task Description}\\
Assemble the LEGO bricks step by step. You can set aside unused pieces on one side and already assembled pieces on the other. The model generates a sequence of frames (or a video) showing its reasoning and assembly process, where LEGO blocks are gradually built.

\textbf{Input Information Structure}\\
You will receive images or video in the following order:
\begin{enumerate}[itemsep=0pt, topsep=0pt]
  \item \textbf{First Image (Input Image)}: The initial image showing the unassembled LEGO pieces scattered or arranged separately.

  \item \textbf{Middle Images or Frames (Model Output - PRIMARY EVALUATION TARGET)}: A sequence of frames or a video showing the model's step-by-step reasoning process as it assembles the LEGO blocks. \textbf{This is what you should evaluate.}

  \item \textbf{Last Image (Ground Truth Reference)}: The correct complete image showing the final assembled LEGO structure. This is provided for comparison reference only, \textbf{NOT for evaluation}.
\end{enumerate}

\textbf{CRITICAL}: Evaluate ONLY the model's process frames (middle images/video), NOT the input or ground truth.

\textbf{Reference Information}
\begin{itemize}[itemsep=0pt, topsep=0pt]
  \item \textbf{Instruction}: ``For this LEGO set, place the orange pieces on the bottom as the first layer, then the gray pieces on the second layer, the yellow pieces on the third layer, and the black pieces on the top layer.''
  \item \textbf{Ground Truth}: The last image shows the correct complete layout for comparison.
\end{itemize}

\textbf{Evaluation Dimensions}\\
Evaluate the overall reasoning process along three percentage-based dimensions (0--100):

\textbf{Background Consistency (0--100)}: In what percentage of frames does the global color, lighting, and visual continuity remain stable? Frames with visual noise, artifact introduction, or block color mismatch reduce the score.

\textbf{Rule Obey (0--100)}: In what percentage of frames does the model follow building rules? Each frame should show valid assembly steps (proper layer ordering, correct block placement, no overlaps or floating pieces).

\textbf{Beneficial Action (0--100)}: In what percentage of frames do the actions effectively build the blocks toward completion? Frames that position or correct blocks toward their ground truth location are beneficial.

\textbf{Visual Quality (0--100)}: In what percentage of frames is the visual quality maintained at a high standard? Frames should be clear, sharp, free from visual noise, artifacts, blurriness, or distortions. Image content should be easily recognizable and distinguishable. Frames with quality degradation, noise, artifacts, blur, or unclear content reduce this score.

\textbf{Output Format}\\
Provide your evaluation strictly in the following JSON format:
\begin{quote}\ttfamily
\{
\{
 "Background\_Consistency": integer 0--100,\\
 "Rule\_Obey": integer 0--100,\\
 "Beneficial\_Action": integer 0--100,\\
 "Visual\_Quality": integer 0--100,\\
 "Explanation": \{\\
 \ \ "Background": "Explain how stable and visually consistent the background remained.",\\
 \ \ "Rule": "Explain how often the LEGO assembly rules were respected (no overlaps, valid placements).",\\
 \ \ "Beneficial": "Explain how frequently the model made progress toward the correct GT arrangement.",\\
 \ \ "Visual\_Quality": "Describe the overall visual quality across frames (clarity, sharpness, absence of artifacts/noise/blur)."\\
 \}\\
\}
\}
\end{quote}

\textbf{Evaluation Principles}
\begin{itemize}[itemsep=0pt, topsep=0pt]
  \item Percentages must be integers (e.g., 90 for 90\%).
  \item Evaluate the full sequence, not only the last few frames.
  \item High scores require consistent, rule-abiding, and progressive assembly behavior.
  \item Visual comparison with the ground truth should guide all judgments.
\end{itemize}

\end{tcolorbox}

\caption{Symbolic Reasoning - Block Building CoT Template.}
\vspace{-0.4cm}
\label{tab:Blocks_Builder_CoT_Template}
\end{table*}

%%%%%%%%%%%%%%%%%%%%%%%%%%%%
% Symbolic Reasoning Block Building Binary Prompt
\begin{table*}[t]
\centering
\small

\begin{tcolorbox}[
  enhanced,
  unbreakable,
  colback=gray!5,
  colframe=green!50!black,
  title=Symbolic Reasoning - Block Building Binary Template
]

You are an expert visual reasoning evaluator for LEGO blocks processes.

\textbf{Task Description}\\
Assemble the LEGO bricks step by step. You can set aside unused pieces on one side and already assembled pieces on the other.

\textbf{Input Information Structure}\\
You will receive images in the following order:
\begin{enumerate}[itemsep=0pt, topsep=0pt]
  \item \textbf{Input Image (First given image)}: The initial image showing the unassembled LEGO pieces scattered or arranged separately.

  \item \textbf{Model Output (Second given image - PRIMARY EVALUATION TARGET)}: The image produced by the model containing its proposed assembled LEGO structure. \textbf{This is what you should evaluate.}

  \item \textbf{Ground Truth Image (Third given image, if provided)}: The correct complete image showing the final assembled LEGO structure. This is provided for comparison reference only, \textbf{NOT for evaluation}.
\end{enumerate}

\textbf{CRITICAL}: Evaluate ONLY the model's output image (second image), NOT the input or ground truth.

\textbf{Reference Information}
\begin{itemize}[itemsep=0pt, topsep=0pt]
  \item \textbf{Instruction}: ``For this LEGO set, place the orange pieces on the bottom as the first layer, then the gray pieces on the second layer, the yellow pieces on the third layer, and the black pieces on the top layer.''
  \item \textbf{Ground Truth}: The third image shows the correct complete layout for comparison.
\end{itemize}

\textbf{Evaluation Dimensions}\\
Evaluate the model output along three binary dimensions (0 or 1):

\textbf{Background Consistency (0 or 1)}: Does the global color, lighting, and visual continuity remain stable? Visual noise, artifact introduction, or block color mismatch should result in 0.

\textbf{Rule Obey (0 or 1)}: Does the model output follow LEGO building rules? The result should show a valid final structure (proper layer ordering, correct block placement, no overlaps or floating pieces).

\textbf{Reasoning Accuracy (0 or 1)}: Is the final solution correct and fully consistent with the ground truth? Blocks must be arranged in the correct layers (orange bottom, gray second, yellow third, black top).

\textbf{Visual Quality (0 or 1)}: Does the model output maintain high visual quality without artifacts or degradation? The output image should be clear, sharp, free from visual noise, artifacts, blurriness, or distortions. Image content should be easily recognizable and distinguishable. Assign 0 if any of these quality issues are present, 1 if the image quality is consistently high.

\textbf{Output Format}\\
Provide your evaluation strictly in the following JSON format:
\begin{quote}\ttfamily
\{
\{
 "Background\_Consistency": 0 or 1,\\
 "Rule\_Obey": 0 or 1,\\
 "Reasoning\_Accuracy": 0 or 1,\\
 "Visual\_Quality": 0 or 1,\\
 "Explanation": \{\\
 \ \ "Background": "Describe whether the visual layout was preserved.",\\
 \ \ "Rule": "Describe if the output follows valid LEGO building rules.",\\
 \ \ "Accuracy": "Describe whether the final result matches the correct solution.",\\
 \ \ "Visual\_Quality": "Describe the overall visual quality of the output image (clarity, sharpness, absence of artifacts/noise/blur)."\\
 \}\\
\}
\}
\end{quote}

\textbf{Evaluation Principles}
\begin{itemize}[itemsep=0pt, topsep=0pt]
  \item Assign ``1'' only if the criterion is fully satisfied without mistakes.
  \item Visual comparison should ensure no loss of context or formatting corruption.
  \item For partial correctness, only \textbf{Rule Obey} may receive 1; all other dimensions must be 0.
\end{itemize}

\end{tcolorbox}

\caption{Symbolic Reasoning - Block Building Binary Template.}
\vspace{-0.4cm}
\label{tab:Blocks_Builder_Binary_Template}
\end{table*}

%%%%%%%%%%%%%%%%%%%%%%%%%%%%
% Klotski Puzzle CoT Prompt
\begin{table*}[t]
\centering
\small

\begin{tcolorbox}[
  enhanced,
  unbreakable,
  colback=gray!5,
  colframe=green!50!black,
  title=Symbolic Reasoning - Klotski Puzzle CoT Template
]

You are an expert visual reasoning evaluator for Klotski Puzzle processes.

\textbf{Task Description}\\
To play the game of Digital Klotski Puzzle, by translating the number squares several times, only one number square can be moved to an empty space each time, so that the final numbers from the upper left to the lower right are arranged in order from 1 to 8, and the lower right corner is empty. The model generates a sequence of frames or a video showing its reasoning and movement process, where number squares are gradually moved.

\textbf{Input Information Structure}\\
You will receive images or video in the following order:
\begin{enumerate}[itemsep=0pt, topsep=0pt]
  \item \textbf{First Image (Input Image)}: The initial klotski puzzle state showing a scrambled arrangement of number squares from 1 to 8 with one empty space.

  \item \textbf{Middle Images or Frames (Model Output - PRIMARY EVALUATION TARGET)}: A sequence of frames or a video showing the model's step-by-step reasoning process as it moves the number squares. \textbf{This is what you should evaluate.}

  \item \textbf{Last Image (Ground Truth Reference)}: The correct complete layout with numbers 1 to 8 arranged in order from the upper left to the lower right and the empty space in the lower right corner. This is provided for comparison reference only, \textbf{NOT for evaluation}.
\end{enumerate}

\textbf{CRITICAL}: Evaluate ONLY the model's process frames (middle images/video), NOT the input or ground truth.

\textbf{Reference Information}
\begin{itemize}[itemsep=0pt, topsep=0pt]
  \item \textbf{Instruction}: ``Only one number square can be moved to an empty space at each step, so that the final numbers from the upper left to the lower right are arranged in order from 1 to 8.''
  \item \textbf{Ground Truth}: The last image shows the correct complete layout for comparison.
\end{itemize}

\textbf{Evaluation Dimensions}\\
Evaluate the overall reasoning process along the following percentage-based dimensions (0--100):

\textbf{Background Consistency (0--100)}: The percentage of frames in which global color, lighting, and visual continuity remain stable.

\textbf{Rule Obey (0--100)}: The percentage of frames in which the model follows Klotski Puzzle movement rules, with exactly one square moved into the empty space at each step and no overlaps or deletions.

\textbf{Beneficial Action (0--100)}: The percentage of frames in which actions effectively move the number squares toward the correct final configuration.

\textbf{Visual Quality (0--100)}: The percentage of frames that maintain high visual quality, remaining clear, sharp, and free from artifacts, noise, blurriness, or distortions.

\textbf{Output Format}\\
Provide your evaluation strictly in the following JSON format:
\begin{quote}\ttfamily
\{
 "Background\_Consistency": integer 0--100,\\
 "Rule\_Obey": integer 0--100,\\
 "Beneficial\_Action": integer 0--100,\\
 "Visual\_Quality": integer 0--100,\\
 "Explanation": \{\\
 \ \ "Background": "Explain how stable and visually consistent the background remained.",\\
 \ \ "Rule": "Explain how often the square movement rules were respected.",\\
 \ \ "Beneficial": "Explain how frequently the model made progress toward the correct ground truth arrangement.",\\
 \ \ "Visual\_Quality": "Describe the overall visual quality across frames (clarity, sharpness, absence of artifacts, noise, or blur)."\\
 \}\\
\}
\end{quote}

\textbf{Evaluation Principles}
\begin{itemize}[itemsep=0pt, topsep=0pt]
  \item Percentages must be integers.
  \item Evaluate the full sequence, not only the final frames.
  \item High scores require consistent, rule-abiding, and progressive movement behavior.
  \item Judgments should be guided by visual comparison with the ground truth.
\end{itemize}

\end{tcolorbox}

\caption{Symbolic Reasoning - Klotski Puzzle CoT Template.}
\vspace{-0.4cm}
\label{tab:Huarong_Road_CoT_Template}
\end{table*}

%%%%%%%%%%%%%%%%%%%%%%%%%%%%
% Klotski Puzzle Binary Prompt
\begin{table*}[t]
\centering
\small

\begin{tcolorbox}[
  enhanced,
  unbreakable,
  colback=gray!5,
  colframe=green!50!black,
  title=Symbolic Reasoning - Klotski Puzzle Binary Template
]

You are an expert visual reasoning evaluator for Klotski Puzzle processes.

\textbf{Task Description}\\
To play the game of Digital Klotski Puzzle, by translating the number squares several times, only one number square can be moved to an empty space each time, so that the final numbers from the upper left to the lower right are arranged in order from 1 to 8, and the lower right corner is empty.

\textbf{Input Information Structure}\\
You will receive images in the following order:
\begin{enumerate}[itemsep=0pt, topsep=0pt]
  \item \textbf{Input Image (First given image)}: The initial klotski puzzle state showing a scrambled arrangement of number squares from 1 to 8 with one empty space.

  \item \textbf{Model Output (Second given image - PRIMARY EVALUATION TARGET)}: The image produced by the model containing its proposed solution. \textbf{This is what you should evaluate.}

  \item \textbf{Ground Truth Image (Third given image, Optional)}: The correct complete image showing the final layout with numbers 1 to 8 arranged in order from the upper left to the lower right. This is provided for comparison reference only, \textbf{NOT for evaluation}.
\end{enumerate}

\textbf{CRITICAL}: Evaluate ONLY the model's output image (second image), NOT the input or ground truth.

\textbf{Reference Information}
\begin{itemize}[itemsep=0pt, topsep=0pt]
  \item \textbf{Instruction}: ``Only one number square can be moved to an empty space each time, so that the final numbers from the upper left to the lower right are arranged in order from 1 to 8.''
  \item \textbf{Ground Truth}: The third image shows the correct complete layout for comparison.
\end{itemize}

\textbf{Evaluation Dimensions}\\
Evaluate the model output along the following binary dimensions (0 or 1):

\textbf{Background Consistency (0 or 1)}: Whether the global color, lighting, and visual continuity remain stable without visual noise, artifacts, or tile color mismatch.

\textbf{Rule Obey (0 or 1)}: Whether the output follows valid Klotski Puzzle movement rules, with exactly one number square moved into the empty space at each step and a proper final grid layout without overlaps or deletions.

\textbf{Reasoning Accuracy (0 or 1)}: Whether the final state matches the ground truth, with numbers 1 to 8 arranged in order from the upper left to the lower right and the empty space in the lower right corner.

\textbf{Visual Quality (0 or 1)}: Whether the output image maintains high visual quality, remaining clear, sharp, and free from artifacts, noise, blurriness, or distortions.

\textbf{Output Format}\\
Provide your evaluation strictly in the following JSON format:
\begin{quote}\ttfamily
\{
 "Background\_Consistency": 0 or 1,\\
 "Rule\_Obey": 0 or 1,\\
 "Reasoning\_Accuracy": 0 or 1,\\
 "Visual\_Quality": 0 or 1,\\
 "Explanation": \{\\
 \ \ "Background": "Describe whether the visual layout was preserved.",\\
 \ \ "Rule": "Describe whether the output follows valid Klotski Puzzle rules.",\\
 \ \ "Accuracy": "Describe whether the final result matches the correct solution.",\\
 \ \ "Visual\_Quality": "Describe the overall visual quality of the output image (clarity, sharpness, absence of artifacts, noise, or blur)."\\
 \}\\
\}
\end{quote}

\textbf{Evaluation Principles}
\begin{itemize}[itemsep=0pt, topsep=0pt]
  \item Assign ``1'' only if the criterion is fully satisfied without mistakes.
  \item Visual comparison should ensure no loss of context or formatting corruption.
  \item For partial correctness, only Rule Obey may receive 1; all other dimensions must be 0.
\end{itemize}

\end{tcolorbox}

\caption{Symbolic Reasoning - Klotski Puzzle Binary Template.}
\vspace{-0.4cm}
\label{tab:Huarong_Road_Binary_Template}
\end{table*}

%%%%%%%%%%%%%%%%%%%%%%%%%%%%
% Maze Navigation Binary Prompt
\begin{table*}[t]
\centering
\small

\begin{tcolorbox}[
  enhanced,
  unbreakable,
  colback=gray!5,
  colframe=green!50!black,
  title=Symbolic Reasoning - Maze Navigation Binary Template
]

You are a visual reasoning evaluator specializing in 2D maze navigation tasks.

\textbf{Task Description}\\
The goal of this task is to navigate a 2D maze starting from the green letter ``S'' and ending at the red letter ``E''. The model must draw a red path that successfully connects ``S'' to ``E'' without crossing any maze walls or altering the maze layout.

\textbf{Input Information Structure}\\
You will receive images in the following order:
\begin{enumerate}[itemsep=0pt, topsep=0pt]
  \item \textbf{Input Image (First given image)}: The original maze with no path drawn, showing only the maze structure with ``S'' (start) and ``E'' (end) markers.

  \item \textbf{Model Output (Second given image - PRIMARY EVALUATION TARGET)}: The maze image generated by the model with its proposed red path. \textbf{This is what you should evaluate.}

  \item \textbf{Ground Truth Image (Third given image, Optional)}: The correct final maze image showing the reference solution path. This is provided for comparison reference only, \textbf{NOT for evaluation}.
\end{enumerate}

\textbf{CRITICAL}: Evaluate ONLY the model's output image (second image), NOT the input or ground truth.

\textbf{Reference Information}
\begin{itemize}[itemsep=0pt, topsep=0pt]
  \item \textbf{Instruction}: ``Starting from the green letter ``S'' and ending at the red letter ``E'', draw a red path that successfully navigates through the maze without crossing walls.''
  \item \textbf{Ground Truth}: The third image shows the correct solution for comparison.
\end{itemize}

\textbf{Evaluation Dimensions}\\
Evaluate the model output image along the following binary dimensions (0 or 1):

\textbf{Background Consistency (0 or 1)}: Whether the maze structure remains identical to the input image, with no added, removed, or modified walls or boundaries.

\textbf{Rule Obey (0 or 1)}: Whether the red path strictly follows maze rules by staying within open corridors and never crossing or overlapping maze walls.

\textbf{Reasoning Accuracy (0 or 1)}: Whether the red path correctly and continuously connects ``S'' to ``E'' as in the ground truth reference.

\textbf{Visual Quality (0 or 1)}: Whether the output image maintains high visual quality, remaining clear, sharp, and free from artifacts, noise, blurriness, or distortions.

\textbf{Output Format}\\
Provide your evaluation strictly in the following JSON format:
\begin{quote}\ttfamily
\{
 "Background\_Consistency": 0 or 1,\\
 "Rule\_Obey": 0 or 1,\\
 "Reasoning\_Accuracy": 0 or 1,\\
 "Visual\_Quality": 0 or 1,\\
 "Explanation": \{\\
 \ \ "Background": "Briefly describe whether the maze structure was preserved.",\\
 \ \ "Rule": "Explain whether the path followed valid maze corridors without crossing walls.",\\
 \ \ "Accuracy": "Explain whether the path correctly connected S to E compared to the ground truth.",\\
 \ \ "Visual\_Quality": "Describe the overall visual quality of the output image (clarity, sharpness, absence of artifacts, noise, or blur)."\\
 \}\\
\}
\end{quote}

\textbf{Evaluation Principles}
\begin{itemize}[itemsep=0pt, topsep=0pt]
  \item Assign ``1'' only if the criterion is fully satisfied without any errors.
  \item Be strict: partial correctness, such as a path that does not reach ``E'', must be scored as 0.
  \item Explanations should be concise and factual.
\end{itemize}

\end{tcolorbox}

\caption{Symbolic Reasoning - Maze Navigation Binary Template.}
\vspace{-0.4cm}
\label{tab:Maze_Navigation_Binary_Template}
\end{table*}

% Maze Navigation CoT Prompt
\begin{table*}[t]
\centering
\small

\begin{tcolorbox}[
  enhanced,
  unbreakable,
  colback=gray!5,
  colframe=green!50!black,
  title=Maze Navigation CoT Template
]

You are an expert visual reasoning evaluator for multi-step 2D maze-solving processes.

\textbf{Task Description}

The goal of this task is to navigate a 2D maze starting from the green letter ``S'' and ending at the red letter ``E''. The model produces a sequence of frames (or a video) showing how it progressively draws the red path through the maze.

\textbf{Input Information Structure}

You will receive images/video in the following order:

\begin{enumerate}[itemsep=0pt, topsep=0pt]
  \item \textbf{First Image (Input Image)}: The original maze with no path drawn, showing only the maze structure with ``S'' (start) and ``E'' (end) markers.

  \item \textbf{Middle Images/Frames (Model Output - PRIMARY EVALUATION TARGET)}: A sequence of frames or video showing the model's step-by-step reasoning process as it draws the red path. \textbf{This is what you should evaluate.}
  \begin{itemize}[itemsep=0pt, topsep=0pt]
    \item \textbf{Important}: ALL images between the first image and the optional last ground truth image are part of the model's output.
    \item If a ground truth image is provided, it will appear AFTER all model frames as the very last image.
    \item If NO ground truth image is provided, ALL images after the first image are model output frames.
  \end{itemize}

  \item \textbf{Last Image (Ground Truth Reference - OPTIONAL)}: If provided, this will be the very last image in the sequence, appearing AFTER all model output frames. The correct final maze image showing the complete correct path from ``S'' to ``E''. This is provided for comparison reference only, NOT for evaluation.
  \begin{itemize}[itemsep=0pt, topsep=0pt]
    \item \textbf{Note}: When no ground truth image is provided, evaluate based on the instruction and maze navigation rules.
  \end{itemize}
\end{enumerate}

\textbf{CRITICAL}: Evaluate ONLY the model's process frames (middle images/video), NOT the input or the optional ground truth reference image.

\textbf{Reference Information}
\begin{itemize}[itemsep=0pt, topsep=0pt]
  \item \textbf{Instruction}: ``Starting from the green letter `S' and ending at the red letter `E', draw a red path that successfully navigates through the maze without crossing walls.''
  \item \textbf{Ground Truth}: The last image shows the correct solution for comparison.
\end{itemize}

\textbf{Evaluation Dimensions}

Assess the model's reasoning process (middle frames) along three percentage-based dimensions (0--100):

\begin{itemize}[itemsep=0pt, topsep=0pt]
  \item \textbf{Background Consistency (0--100)}: In what percentage of the model's frames does the maze structure remain unchanged? (Any alteration of walls, boundaries, or maze lines in the model's output decreases this score.)

  \item \textbf{Rule Obey (0--100)}: In what percentage of the model's frames does the red path obey maze rules? (Crossing walls, overlapping borders, or jumping across corridors in the model's output reduces this score.)

  \item \textbf{Beneficial Action (0--100)}: In what percentage of the model's frames does the model make progress toward the correct solution? (Steps that extend the correct path toward ``E'' increase this score; random or regressive actions reduce it.)

  \item \textbf{Visual Quality (0--100)}: In what percentage of the model's frames is the visual quality maintained at a high standard? (Frames should be clear, sharp, free from visual noise, artifacts, blurriness, or distortions. Image content should be easily recognizable and distinguishable. Frames with quality degradation, noise, artifacts, blur, or unclear content reduce this score.)
\end{itemize}

\textbf{Output Format}

Return your evaluation in the following JSON format:

\begin{quote}\ttfamily
\{\{
"Background\_Consistency": integer 0--100,
"Rule\_Obey": integer 0--100,
"Beneficial\_Action": integer 0--100,
"Visual\_Quality": integer 0--100,
"Explanation": \{
"Background": "Explain how often the maze layout stayed intact across the model's frames.",
"Rule": "Explain how frequently maze rules were followed during the model's path drawing.",
"Beneficial": "Describe whether the model's actions generally moved closer to solving the maze.",
"Visual\_Quality": "Describe the overall visual quality across frames (clarity, sharpness, absence of artifacts/noise/blur)."
\}
\}\}
\end{quote}

\textbf{Evaluation Principles}
\begin{itemize}[itemsep=0pt, topsep=0pt]
  \item Percentages should be integers (e.g., 85 for 85\%).
  \item Evaluate the full temporal sequence of the MODEL'S OUTPUT — consider overall trends, not single-frame anomalies.
  \item A high Beneficial Action score requires consistent, goal-directed steps toward ``E'' in the model's process.
  \item Compare the model's process against the ground truth reference (last image) to assess correctness.
  \item Provide concise, factual explanations for each dimension.
\end{itemize}

\end{tcolorbox}

\caption{Maze Navigation CoT Template.}
\vspace{-0.4cm}
\label{tab:Maze_Navigation_CoT_Template}
\end{table*}

\end{document}